\documentclass[letterpaper, 10 pt, conference]{ieeeconf}  % Comment this line out if you need a4paper

\IEEEoverridecommandlockouts                              % This command is only needed if
\usepackage{times}
\usepackage{soul}
\usepackage{url}
\usepackage[hidelinks]{hyperref}
\usepackage[utf8]{inputenc}
\usepackage[small]{caption}
\usepackage{graphicx}
\usepackage{amsmath,amssymb}

\usepackage{amsthm}
\usepackage{booktabs}
\newtheorem{theorem}{Theorem}

\usepackage{algpseudocode,algorithm}
\usepackage[subrefformat=parens]{subcaption}
\usepackage{amssymb}
\title{\LARGE \bf
TAdam: A Robust Stochastic Gradient Optimizer
}

\author{Wendyam Eric Lionel Ilboudo$^1$, Taisuke Kobayashi$^{1}$, and Kenji Sugimoto$^{1}$% <-this % stops a space
\thanks{$^{1}$All authors are with the Division of Information Science, Nara Institute of Science and Technology, 8916-5 Takayama-cho, Ikoma, Nara 630-0192, Japan
{\tt\small \{ilboudo.wendyam\_eric.in1, kobayashi, kenji\}@is.naist.jp}%
}%
}

\begin{document}

\maketitle
\thispagestyle{empty}
\pagestyle{empty}

%%%%%%%%%%%%%%%%%%%%%%%%%%%%%%%%%%%%%%%%%%%%%%%%%%%%%%%%%%%%%%%%%%%%%%%%%%%%%%%%
\begin{abstract}
    Machine learning algorithms aim to find patterns from observations, which may include some noise, especially in robotics domain.
    To perform well even with such noise, we expect them to be able to detect outliers and discard them when needed.
    We therefore propose a new stochastic gradient optimization method, whose robustness is directly built in the algorithm, using the robust student-t distribution as its core idea.
    Adam, the popular optimization method, is modified with our method and the resultant optimizer, so-called TAdam, is shown to effectively outperform Adam in terms of robustness against noise on diverse task, ranging from regression and classification to reinforcement learning problems.
\end{abstract}

%%%%%%%%%%%%%%%%%%%%%%%%%%%%%%%%%%%%%%%%%%%%%%%%%%%%%%%%%%%%%%%%%%%%%%%%%%%%%%%%
\section{Introduction}

% Move 1: Establishing the research area
% intro of SGD and its importance in ML
The field of machine learning is undoubtedly dominated by first-order optimization methods based on the gradient descent algorithm and particularly~\cite{bottou2010large}, its stochastic variant, the stochastic gradient descent (SGD) method~\cite{sgd_paper}.
The popularity of the SGD algorithm comes from its simplicity, its computational efficiency with respect to second-order methods, its applicability to online training and its convergence rate that is independent of the training set.
In addition, SGD has high affinity with deep learning~\cite{deep_learning}, where network parameters are updated by backpropagation of their gradients, and is intensively used to train large deep neural networks.

% problems
% [comment] Here, please focus on robotics domain, where there are many types of noises: observation, annotation, ..., because this paper will be submitted to RA-L.
Despite such established popularity, a specific trait of SGD is the inherent noise, coming from sampling training points.
Even though this stochasticity makes the algorithm more likely to find a global minimum, those fluctuations also slow down the learning process and furthermore, render the algorithm sensitive to outliers.
Indeed, bad estimates of the gradients are likely to produce bad estimation of the minimum.

Many of the new optimizers proposed to improve the SGD algorithm and tackle complex training scenarios where gradient descent methods behave poorly also share the same weakness to aberrant value.
Adam (Adaptive moment estimates)~\cite{adam}, one of the most widely used and practical optimizers for training deep learning models, is no exception.
This is mainly due to the insufficient number of samples implicitly involved in its first moment evaluation.

 The weakness to noisy data is particularly important in robotics learning where incomplete, ambiguous and noisy sensor data are inevitable. Furthermore, in order to generate large scale robot datasets for scaling up robot learning, the ability to use
automatically labeled data~\cite{suchi2019easylabel} is important. Robust learning methods are therefore needed to deal with the eventual noisy labels and can improve the performance of low-cost robots that suffers from inaccurate position control and calibration and noisy executions, without the need of a noise modeling network~\cite{gupta2018robot}.

% State the objectives
Hence, the aim of the present research is to propose a robust version of Adam through the use of robust estimates of the momentum, which is assumed to be the first-order probabilistic moment of the gradients.
The key idea for such robust estimates is the use of the student-t distribution, which is a model suitable for the estimates from a few samples~\cite{prml}.

%%%%%%%%%%%%%%%%%%%%%%%%%%%%%%%%%%%%%%%%%%%%%%%%%%%%%%%%%%%%%%%%%%%%%%%%%%%%%%%%
\section{Background and previous works}

\subsection{Background}

\paragraph{Stochastic Gradient Descent} Let $x_t$ be a random sample from the data set at iteration $t$, $J_\theta(x_t)$ the objective function evaluated on data $x_t$ with the parameters $\theta$, $g_t = \nabla_{\theta}J_\theta(x_t)$ its gradient, and $\alpha$ the learning rate.
The SGD algorithm~\cite{sgd_paper} updates $\theta_{t-1}$ to $\theta_t$ through the following update rule:
\begin{align}
    \theta_t = \theta_{t-1} - \alpha g_t \label{eq:sgd_update}
\end{align}
This algorithm yields at least local minima of $J$ w.r.t $\theta$.

\paragraph{Improving SGD} Since its proposition, many ideas have been developed in order to improve the convergence property of the SGD algorithm.
This feature heavily connects to the fluctuations of the gradients during learning.
All the research that aim to accelerate the convergence rate have done so through several approaches.
For instance, they improved i) the update method of the parameters~\cite{sgd_momentum,nag_paper,sag_paper,svrg_paper}; ii) the adjustment of the learning rate~\cite{adagrad,adadelta,rmsprop,adabound}; and iii) the robustness to aberrant values from heavy-tailed data~\cite{vsgd_fd,adasecant,roadam,holland2019efficient}.
Those approaches have culminated to some pretty effective state-of-the-art first-order optimization methods, going from the momentum idea to the adaptive learning rate and variance reduction schemes.
Below, we review some of the works related to the robustness.

\subsection{Previous works}

As stated before, SGD is inherently noisy and susceptible to produce bad minima estimates when facing aberrant gradient estimates.
A lot of work have therefore been done to propose more robust methods for efficient machine learning under noise or data with heavy tails.

In this review, we ignore the general statistical methods for robust mean estimates~\cite{robustmeanestim} such as the median based estimations~\cite{geometricmedian,riskmedian,lossmedian} due to their practical limitations.
Three main approaches are distinguished: a) methods based on direct robust estimates of the loss (or risk) function~\cite{mestim_riskminimization}; b) methods based on robust estimates of the gradients~\cite{byzantineGD,holland2019efficient} among which falls our algorithm; and c) methods with small learning rates for wrong gradient estimates~\cite{roadam}.

\paragraph{Robust risk estimation} Those methods usually require the use of all the available data in order to produce, for each parameter, a robust estimate of the loss function to be minimized. A specific inconvenient trait of this approach is the implicit definition of the robust estimate, which may introduce some computational roadblocks.
As briefly explained by Holland et al.~\cite{holland2019efficient}, since the estimates do not need to be convex even in the case where the loss function is, the non-linear optimization can be both unstable and costly in high dimensions.

\paragraph{Robust gradient descent} This approach usually rely on the replacement of the empirical mean (first moment) gradient estimate with a more robust alternative, and simply differs in the method used to achieve this objective.
Chen et al.~\cite{byzantineGD} proposed the use of the geometric median of the gradients mean to aggregate multiple candidates. Using the same strategy, Prasad et al.~\cite{prasadrobustGD} proposed a class of gradient estimator based on the idea that the gradient of a population loss could be regarded as the mean of a multivariate distribution, reducing the problem of gradient estimation to a multivariate mean estimation problem.
Very close to our approach, Holland et al.~\cite{holland2019efficient} proposed to carefully reduce the effect of aberrant values instead of discarding them, which can also result in unfortunate discards of valuable data.

\paragraph{Adaptive learning rate} This approach is to reduce the effect of wrong gradient estimates by reducing the learning rate.
One such approach has been proposed by Haimin et al.~\cite{roadam} and shares the same objective as ours to produce a robust version of the Adam optimization algorithm.
The method employed by Haimin et al. uses an exponential moving average (EMA) of the ratio between the current loss value $l_{t}$ and the past one $l_{t-1}$ to scale the learning rate.
However, this strategy allows the outliers to modify the estimated gradient mean, and then uses the impact of the deviated mean on the loss function to reduce the effect on subsequent updates.

As one of the problems in the EMA scheme, the lack of robustness has been dealt with in~\cite{vsgd_fd} and~\cite{adasecant}.
In those methods, the exponential decay parameter of the EMA is increased whenever a value that falls beyond some boundary is encountered.
The common drawback in this strategy is that all outlier gradients are treated equally and discretely without consideration of how far they are from the normal values, and the boundary over which a data is considered to be an outlier must be set manually before training.

\paragraph{Our contribution} To the best of our knowledge, our approach, named TAdam, is the first to employ estimates of the student-t distribution first moment to replace the estimates of the Gaussian first moment introduced by Adam, through the EMA scheme.
The main advantage of this approach is that it relies on the natural robustness of the student-t distribution and its ability to deal with outliers, and can easily be reduced to Adam for non-heavy-tailed data.
Also, even though, in this letter, we use our method to modify the popular optimizer Adam, we encourage the reader to keep in mind that it can be integrated to the other stochastic gradient descent methods that rely on EMAs like RMSProp~\cite{rmsprop}, VSGD-fd~\cite{vsgd_fd}, Adasecant~\cite{adasecant} or Adabound~\cite{adabound}.
% [comment] please refer other EMA-based SGDs

%%%%%%%%%%%%%%%%%%%%%%%%%%%%%%%%%%%%%%%%%%%%%%%%%%%%%%%%%%%%%%%%%%%%%%%%%%%%%%%%
\section{Proposal}
%%%%%%%%%%%%%%%%%%%%%%%%%%%%%%%%%%%%%%%%
\subsection{Adaptive moment estimation: Adam}

Before describing our proposal, let us introduce Adam~\cite{adam}, the baseline of TAdam.
Adam is a popular method that combines the advantages of SGD with momentum along with those of adaptive learning rate methods~\cite{rmsprop,adadelta}.
Its update rule is implemented as follows:
\begin{align}
    m_t &= \beta_1 m_{t-1} + (1 - \beta_1) g_t \label{eq:adam_m}\\
    v_t &= \beta_2 v_{t-1} + (1 - \beta_2) {g_t}^2 \label{eq:adam_v}\\
    \theta_t &= \theta_{t-1} - \alpha \frac{m_t}{(1 - {\beta}^t_1)(\sqrt{v_t(1 - {\beta}^t_2)^{-1}} + \epsilon)} \label{eq:adam_update}
\end{align}
where $m_t$ is the first-order moment (i.e., mean of gradients) and $v_t$ is the second-order moment utilized to adjust learning rates at time step $t$.
$\beta_1$ and $\beta_2$ are the exponential decay rates (by default $0.9$ and $0.999$, respectively).
$\alpha$ is the global learning rate and $\epsilon$ is a small value added to avoid division by zero (typical value of $10^{-8}$).

\begin{figure}[tb]
    \centering
    \includegraphics[keepaspectratio=true,width=0.6\linewidth]{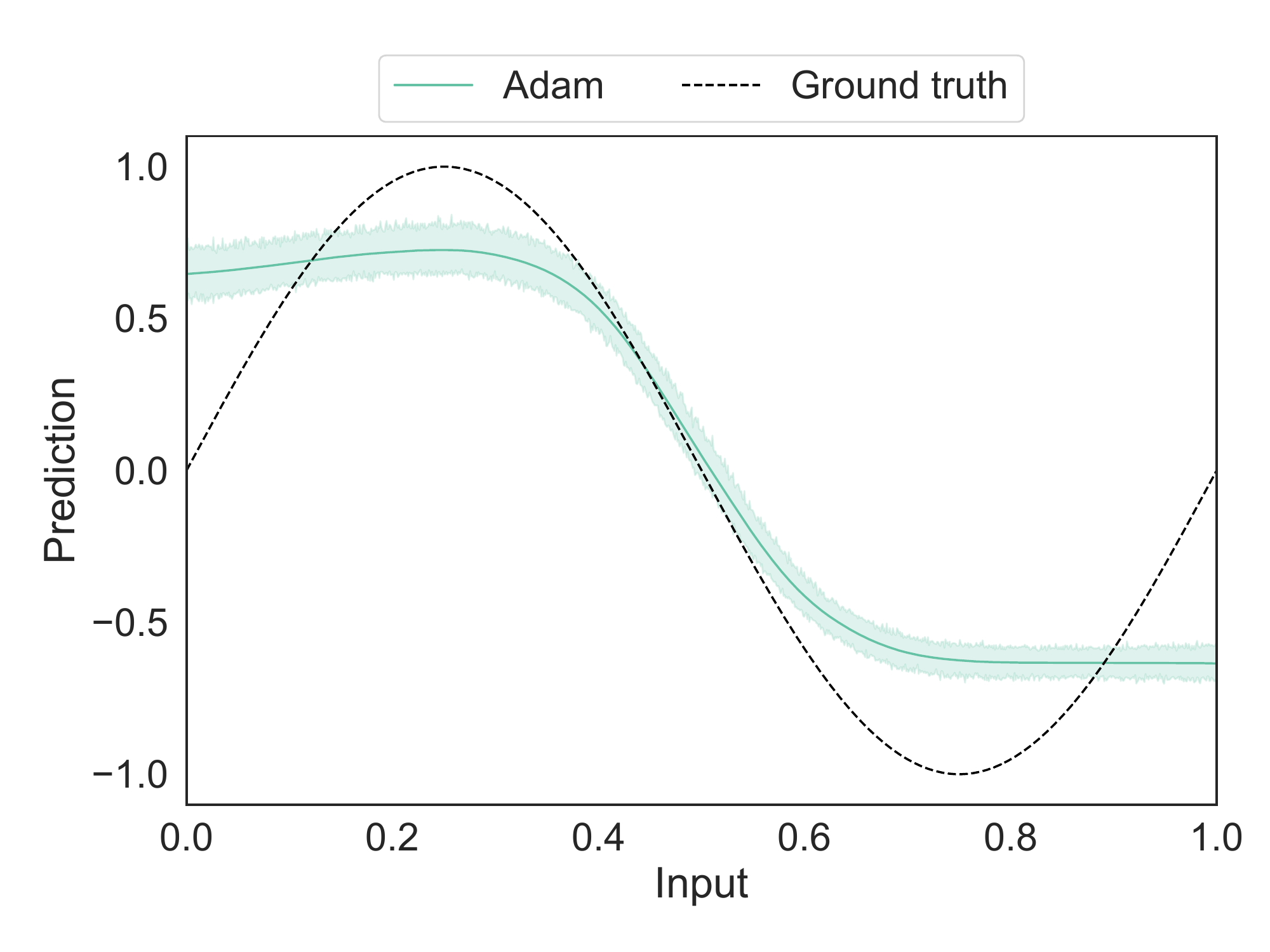}
    \caption{Sensitivity to outliers in Adam:
    a regression task with noise drawn from a student-t distribution with degrees of freedom $\nu = 1$, $\mu = 0$ and scale $\lambda = 0.05$ was conducted with Adam;
    the predicted curve had large variance and its accuracy was clearly deteriorated due to the noises.
    }
    \label{fig:adam_drawback}
\end{figure}

Although the use of EMAs in equations~\eqref{eq:adam_m} and \eqref{eq:adam_v} makes the gradients smooth and reduces the fluctuations inherent to SGD, they are also sensitive to outliers.
In particular, with a small value like $\beta_1$ ($=0.9$), the momentum $m_t$ is very likely to be pulled out by outliers and easily deviate from the true average.
This fluctuation makes learning unstable (see Fig.~\ref{fig:adam_drawback}), and therefore, more robust learning techniques are needed.

%Figure 1

%%%%%%%%%%%%%%%%%%%%%%%%%%%%%%%%%%%%%%%%
\subsection{Overview}

Our proposition relies on the fact that the EMA, like equations~\eqref{eq:adam_m} and \eqref{eq:adam_v}, can be regarded as an incremental update law of the mean in the normal distribution with a fixed number of samples.
The sensitivity of Adam to aberrant gradient values is therefore just a feature inherited from the normal distribution, which is itself also sensitive to outliers.

\begin{figure}[tb]
    \centering
    \includegraphics[keepaspectratio=true,width=0.7\linewidth]{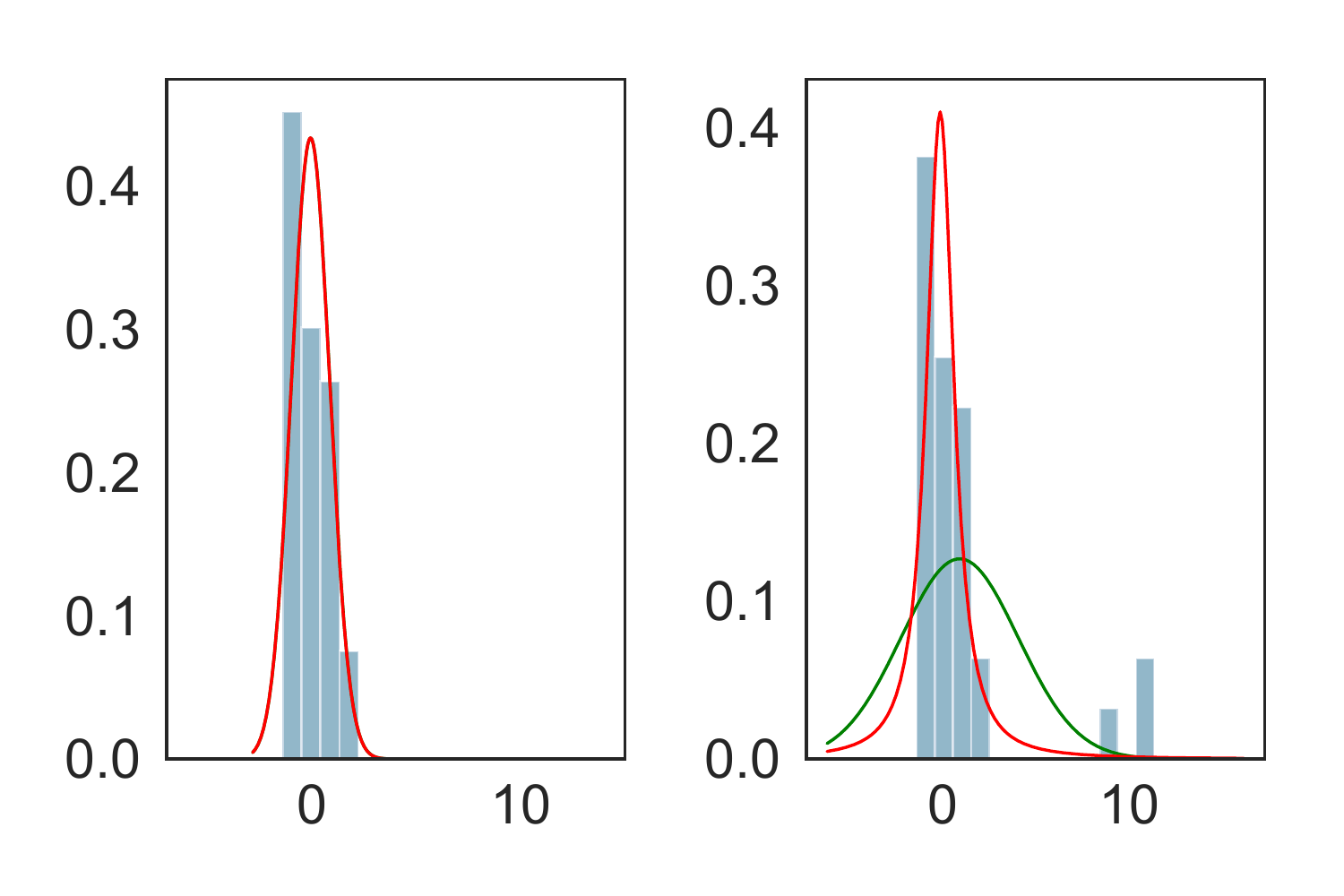}
    \caption{Robustness to outliers:
    the normal distribution (in green) was pulled out by outliers;
    in contrast, the student-t distribution (in red) allowed their existence and hardly moved.
    }
    \label{fig:student_advantage}
\end{figure}

In order for Adam to be robust, the distribution of the gradients must be assumed to come from a robust probability distribution.
We therefore propose to replace the normal distribution moment estimates by those from the student-t distribution, which is well-known to be a robust probability distribution~\cite{prml,robust_student_t1,robust_student_t2}, as shown in Fig.~\ref{fig:student_advantage}, and a general form of the normal distribution.
From the next section, we describe how the EMA is replaced using the student-t distribution, and the features of our implementation are analyzed later.
A pseudo code of TAdam is summarized in Algorithm~\ref{alg:tadam}.

%Figure 2

%Algorithm
\begin{algorithm}[tb]
    \caption{TAdam, our proposed algorithm for stochastic optimization: it is an extended version of Adam by the student-t mean estimation; in typical setting, $\beta_1 = 0.9$ is smaller than $\beta_2=0.999$; a good default value for the degrees of freedom is found, $\nu = d = dim[\nabla_{\theta}f(\theta)]$.}
    \label{alg:tadam}
    \begin{algorithmic}[1]
        % requirements
        \Require{$\alpha$: Learning rate}
        \Require{$\beta_1$, $\beta_2 \in$ [0,1): Exponential decay rates}
        \Require{$\epsilon$: Small term added to the denominator}
        \Require{$\nu$: Degrees of freedom}
        \Require{$J_\theta(x_t)$: Objective function with parameters $\theta$}
        \Require{$\theta_0$: Initial parameters}
        % initialization
        \State{$m_0 \gets 0$, $v_0 \gets 0$, $t \gets 0$}
        \State{$W_0 \gets \frac{\beta_1}{1-\beta_1}$}
        % main loop
        \While{$\theta_t$ not converged or $t<T_\mathrm{max}$}
        \State{$t \gets t+1$}
        \State{$x_t \gets x_{t+1}$}
        \State{$g_t \gets \nabla_{\theta}J_{\theta_{t-1}}(x_t)$}
        \State{$w_t \gets (\nu + d)\left (\nu + \sum\limits_j \frac{(g_t^j - m_{t-1}^j)^2}{v_{t-1}+\epsilon} \right )^{-1}$}
        \State{$m_t \gets \frac{W_{t-1}}{W_{t-1} + w_t}m_{t-1} + \frac{w_t}{W_{t-1} + w_t}g_t$}
        \State{$W_t \gets \frac{2\beta_1 - 1}{\beta_1}W_{t-1} + w_t$}
        \State{$v_t \gets \beta_2 v_{t-1} + (1 - \beta_2) {g_t}^2$}
        \State{$\theta_t \gets \theta_{t-1} - \alpha \frac{m_t}{(1 - {\beta}^t_1)(\sqrt{v_t(1 - {\beta}^t_2)^{-1}} + \epsilon)}$}
        \EndWhile
        % end
        \State{\Return{$\theta_t$}}
    \end{algorithmic}
\end{algorithm}

%%%%%%%%%%%%%%%%%%%%%%%%%%%%%%%%%%%%%%%%
\subsection{Formulation}

To replace the EMA by the student-t distribution, a new hyperparameter, the degrees of freedom of the student-t distribution $\nu$, is introduced to control the robustness.

We can derive the incremental update law of the first moment $\mu$ for the student-t distribution using a maximum log-likelihood estimator.
Given $x_1, \ldots, x_n$ $d$-dimensional i.i.d. random samples from multivariate student-t distribution $p_t$ with the parameters $\mu$, $\Sigma$ and $\nu$, its log-likelihood function is expressed as:
\begin{align}
    \begin{split}
        log(p_t) &= \left\{n\log\Gamma\left(\frac{\nu + d}{2} \right) - n\log\Gamma\left(\frac{\nu}{2}\right) \right. \\
        &\left. - \frac{n\nu}{2}\log (\nu) - \frac{nd}{2}\log (\pi) - \frac{n}{2}\log (|\Sigma|) \right. \\
        &\left. - \left(\frac{\nu + d}{2}\right)\sum_{i=1}^n \log (\nu + D_i) \right\}
    \end{split}
    \label{eq:t_loglikelihood}
\end{align}
where $D_i = (x_i - \mu)^T\Sigma^{-1}(x_i - \mu)$.
Taking the gradient with respect to $\mu$ and setting it equal to $0$ gives us:
\begin{align}
    \frac{\partial log(p_t)}{\partial \mu} &= \sum_{i=1}^n (\nu + d)\frac{x_i - \mu}{\nu + D_i} \nonumber \\
    &= \sum_{i=1}^n x_i\frac{\nu + d}{\nu + D_i} - \mu \sum_{i=1}^n \frac{\nu + d}{\nu + D_i} = 0 \label{eq:t_grad_halfsolved}
\end{align}
If we solve this equation for $\mu$, we get the expression of the first moment estimate given $n$ samples:
\begin{align}
    \hat{\mu}_n &= \frac{\sum_{i=1}^n x_i w_i}{W_n} = \frac{\sum_{i=1}^{n-1} x_i w_i + x_n w_n}{W_{n-1} + w_n} \nonumber \\
    &= \frac{W_{n-1}}{W_{n-1} + w_n} \hat{\mu}_{n-1} + \frac{w_n}{W_{n-1} + w_n}  x_n \label{eq:tAdam_mu}
\end{align}
where $w_i = (\nu + d)/(\nu + D_i)$ and $W_n = \sum_{i=1}^n w_i$.

By assuming a diagonal distribution and fixing the number of samples (decaying $W_n$), we can derive the equation~\eqref{eq:tadam_m} used below in TAdam.
Due to the high value of $\beta_2$ (i.e., $0.999$ about $1000$ samples) w.r.t. $\beta_1$ (i.e., $0.9$ about $10$ samples), only the first-order moment in equation~\eqref{eq:adam_m} is replaced by the following rule:
\begin{align}
    m_t = \frac{W_{t-1}}{W_{t-1} + w_t}m_{t-1} + \frac{w_t}{W_{t-1} + w_t}g_t \label{eq:tadam_m}
\end{align}
where
\begin{align}
    w_t &= \frac{\nu + d}{\nu + \sum_j^d \frac{(g_t^j - m_{t-1}^j)^2}{v_{t-1}^j+\epsilon}} \label{eq:tadam_w}\\
    W_t &= \frac{2\beta_1 - 1}{\beta_1}W_{t-1} + w_t \label{eq:tadam_W}
\end{align}
$v_{t-1}$ is the unmodified Adam's second moment estimate coming from equation~\eqref{eq:adam_v}, and $d$ is the dimension of the gradient $g_t$ (i.e., the number of parameters in subsets like layers of deep learning).
Here, the summation in the denominator of $w_t$ is substituted from now on by $D_t$ since it corresponds to the Mahalanobis distance between the gradient of the parameter $\theta_j$, $g_t^j$, and the corresponding previous estimate of the mean, $m_{t-1}^j$, w.r.t. the variance that is assumed to be the same as Adam's second moment estimate, $v_{t-1}^j$.
Note that, ultimately, the gradients converge to zero, and therefore, the second moment would be consistent with the variance of the gradients.

The power of this update rule is two folds: the outliers detection and the robustness control.
Their details are explained below.

%%%%%%%%%%%%%%%%%%%%%%%%%%%%%%%%%%%%%%%%
\subsection{The outliers detection}

This is performed through $w_t$ which is an adaptive weight of the mean introduced in equation~\eqref{eq:tadam_m} with degrees of freedom $\nu$.
Again, we can notice that $w_t$ depends on the Mahalanobis distance $D_t$.
Hence, outlying gradient values are down-weighted since their Mahalanobis distances are larger than for normal values, and their contribution to the momentum update is therefore automatically dampened.
On the contrary, the normal gradients are up-weighted ultimately by $1 + d/\nu$ due to zero Mahalanobis distances, although $m$ is kept in that case since $m_{t-1} = g_t$.
In short, TAdam automatically and continuously reduces only the adverse effects of the outlier gradients.

%%%%%%%%%%%%%%%%%%%%%%%%%%%%%%%%%%%%%%%%
\subsection{The robustness control} \label{robustness}

The Student-t distribution has a controllable robustness and that nice property of being similar to the normal distribution when the degrees of freedom grows larger.
The same feature is left in TAdam, as can be seen in equation~\eqref{eq:tadam_w}.
Namely, when $\nu \rightarrow \infty$, we have:
\begin{align}
    \lim_{\nu \rightarrow \infty} w_t = \frac{\infty + d}{\infty + D_t} = 1
\end{align}
In this case, TAdam loses its robustness to outliers, like Adam.

To make TAdam be an extended version of Adam, the decay rule in equation~\eqref{eq:tadam_W} is designed to fulfill some requirements.
Specifically, if $\nu \rightarrow \infty$, the decay rate derived from $W_{t-1}$ and $w_t$ in equation~\eqref{eq:tadam_m} must be consistent with $\beta_1$ at any time.
\begin{align}
    W_t = W_0 = \frac{\beta_1}{1-\beta_1},  \forall t > 0
\end{align}
To satisfy such a constant $W$, the decay rate in equation~\eqref{eq:tadam_W} can be derived as follows if the decay rule is given as $W_t \leftarrow \gamma W_{t-1} + w_t$.
\begin{align}
\gamma = \frac{W_t - w_t}{W_{t-1}} = \frac{2\beta_1 - 1}{\beta_1}
\end{align}
By the above derivation, TAdam defined by equations~\eqref{eq:tadam_m}--\eqref{eq:tadam_W} is proved to be the extended version of Adam defined by the equation~\eqref{eq:adam_m} (and equations~\eqref{eq:adam_v}--\eqref{eq:adam_update}).

%%%%%%%%%%%%%%%%%%%%%%%%%%%%%%%%%%%%%%%%
\subsection{The Regret Bound and TAdam's Convergence}

The convergence of the TAdam algorithm is assured by the two following theorems, whose proofs can be found in the appendix:
\begin{theorem}
    Given $\{\theta_t\}_0^T$ and $\{v_t\}_0^T$, the sequences obtained from the TAdam algorithm, $\alpha_t = \frac{\alpha}{\sqrt{t}}$, $\beta_w = \beta_{1t}$, $\mathbb{E}[\beta_w] \leq \bar{\beta}_w < 1$ and $\gamma = \frac{\bar{\beta}_w}{\sqrt{\beta_2}} < 1$.
    If $F$ has a bounded diameter $D_{\infty}$, and if $g = \parallel\nabla f_t(\theta)\parallel_{\infty} \leq G_{\infty}$ for all $t \in [T]$ and $\theta \in F$.
    % [comment] refer AMSGrad
    Then, for $\theta_t$ generated using TAdam (with the AMSGrad~\cite{reddi2019amsgrad} scheme), we have the following upper bound on the regret:
    \begin{align}
        \begin{split}
            R_T \leq &\left.\frac{D^2_{\infty}}{2\alpha_T (1-\bar{\beta}_w)} \sum\limits_{i=1}^d \hat{v}_{T,i}^{1/2} + \frac{D^2_{\infty}}{(1-\bar{\beta}_w)^2} \sum\limits^T_{t=1} \sum\limits_{i=1}^d \frac{\beta_{1t}\hat{v}_{t,i}^{1/2}}{\alpha_t} \right.\\
            &\left. + \frac{\alpha \sqrt{1 + \log T}}{(1-\bar{\beta}_w)^2(1 - \gamma)\sqrt{(1-\beta_2)}} \sum\limits^d_{i=1} \parallel g_{1:T, i}\parallel_2 \right.
        \end{split}
        \label{eq:th_regret_bound}
    \end{align}
\end{theorem}
\begin{theorem}
    % [comment] \mathbb{N} is normal distribution or what?
    Let’s assume that the gradients $g$ ultimately follow an asymptotic Normal distribution $g \in \mathbb{R}^d \sim \mathcal{N}$, according to the central limit theorem; then the Mahalanobis distance appearing in TAdam follows a Chi-Squared distribution $D_M^2(g, \mu) = \sum\limits_j \frac{(g_t^j - \mu_j)^2}{v_j} \sim \chi^2(d)$, and the expected value of the adaptive decay parameter $\beta_w = \frac{W_{t-1}}{W_{t-1} + w_t}$ is constrained, for $\beta_1 < 1$, by the following relation:
    \begin{align}
        \mathbb{E}[\beta_w] &\leq \beta_1
    \end{align}
    \label{eq:th_bw_bound}
\end{theorem}

 We can see that the difference between the upper bound of TAdam and Adam lies in the value of $\bar{\beta}_w$, which corresponds to the expected value of the adaptive exponential decay parameter $\beta_w = \frac{W_{t-1}}{W_{t-1} + w_t}$.
Theorem~\eqref{eq:th_bw_bound} tells us that, if the gradients are normally distributed, this value is bounded above by $\beta_1$, so that we can recover the same upper bound for TAdam and Adam.
 However, if we know the exact value of the expected value, a more precise upper bound for the regret can be obtained.

%%%%%%%%%%%%%%%%%%%%%%%%%%%%%%%%%%%%%%%%%%%%%%%%%%%%%%%%%%%%%%%%%%%%%%%%%%%%%%%%
\section{Experiments}

 To assess the robustness of TAdam against noisy data, we conducted three types of experiments spanning the main machine learning frameworks, i.e. supervised learning (regression and classification) and reinforcement learning.
We compare TAdam mainly with Adam, but also with another robust gradient descent algorithm, RoAdam~\cite{roadam}.
 
 %Figure Regression
\begin{figure*}[tb]
    \centering
    \begin{minipage}[t]{0.20\linewidth}
        \centering
        \includegraphics[keepaspectratio=true,width=\linewidth]{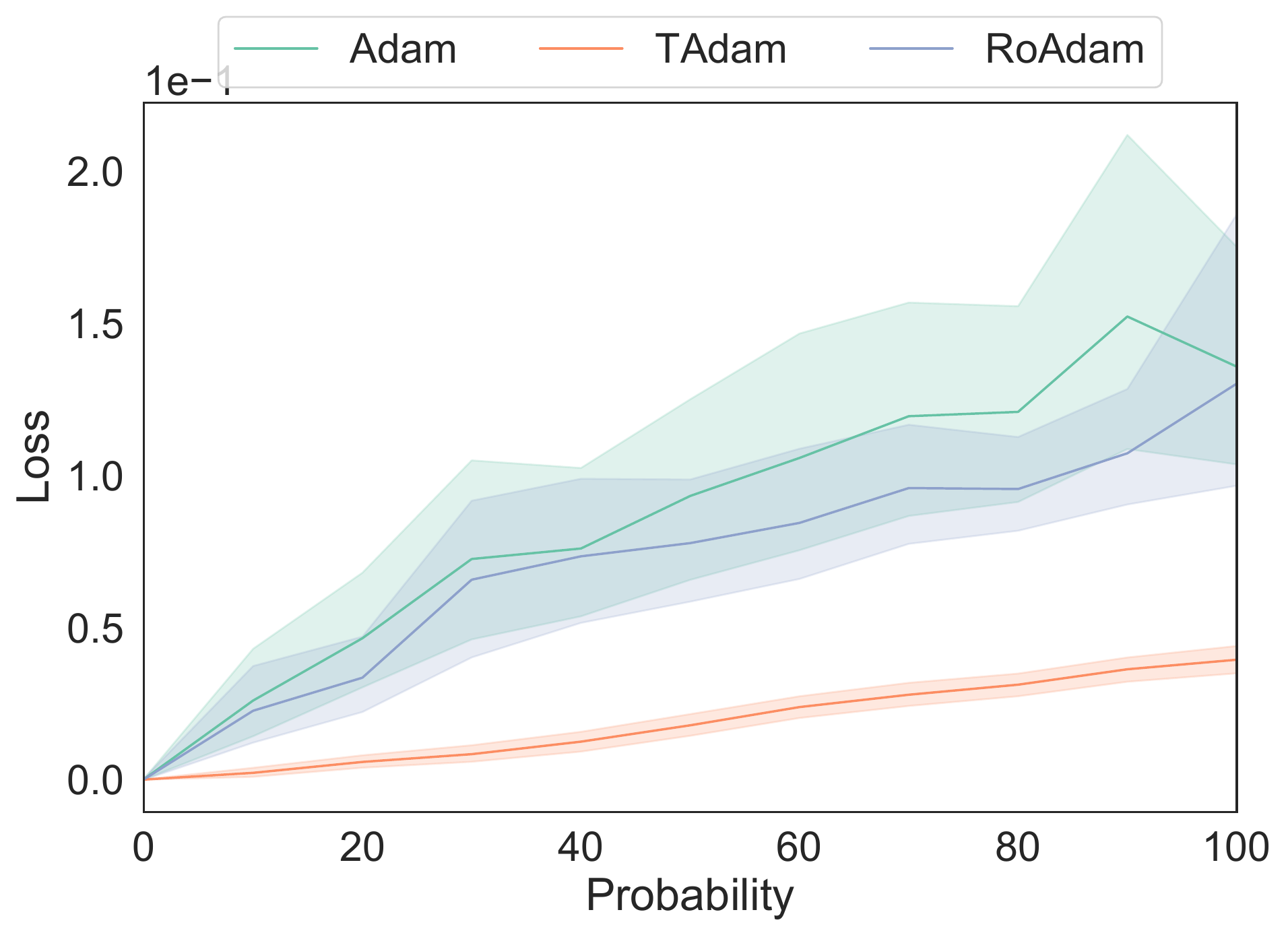}
        \subcaption{$(\nu_{\zeta}, \lambda_{\zeta}) = (1.0, 0.05)$}
        \label{fig:result_lrt_1_005}
    \end{minipage}
    \centering
    \begin{minipage}[t]{0.19\linewidth}
        \centering
        \includegraphics[keepaspectratio=true,width=\linewidth]{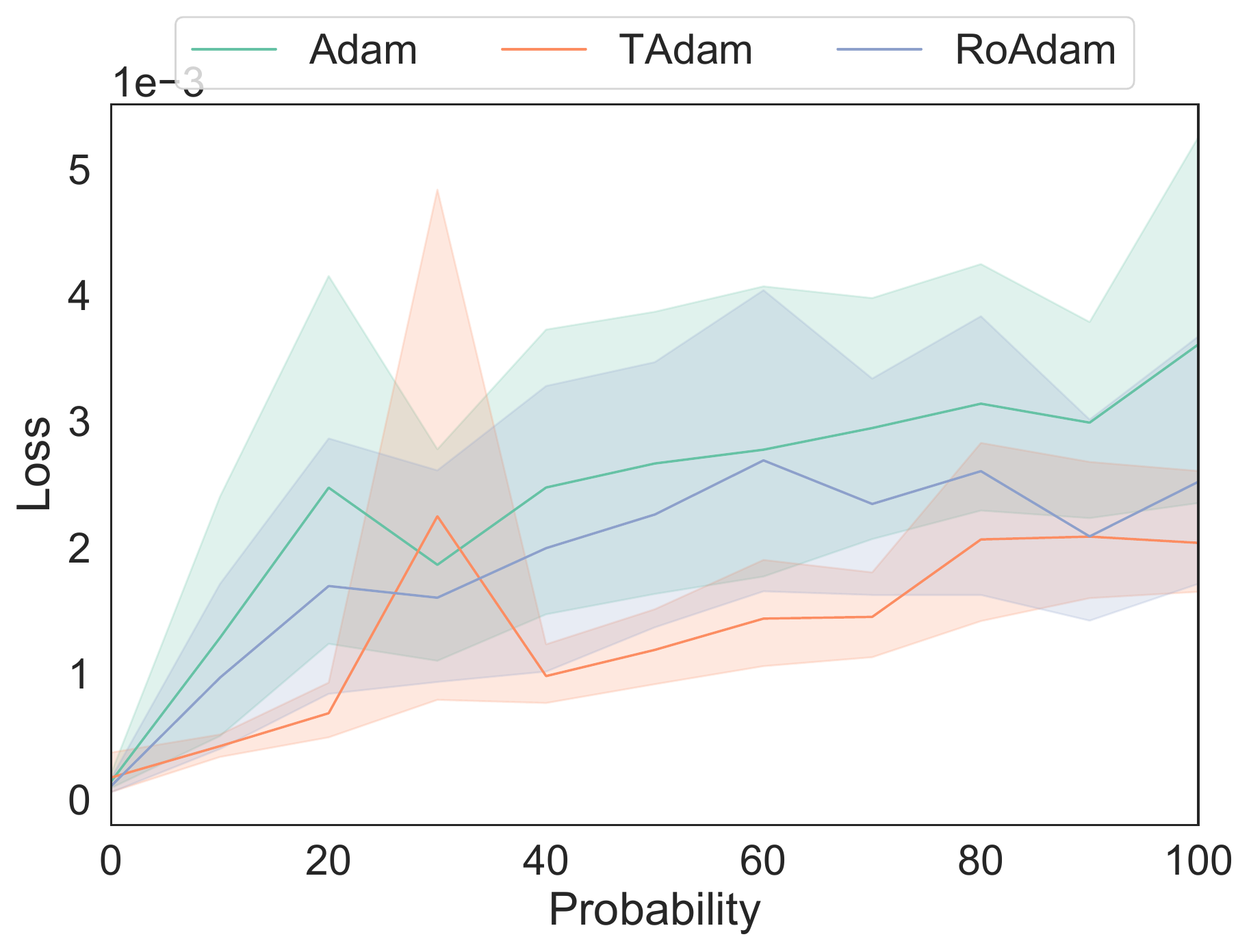}
        \subcaption{$(\nu_{\zeta}, \lambda_{\zeta}) = (2.0, 0.03)$}
        \label{fig:result_lrt_2_003}
    \end{minipage}
    \centering
    \begin{minipage}[t]{0.20\linewidth}
        \centering
        \includegraphics[keepaspectratio=true,width=\linewidth]{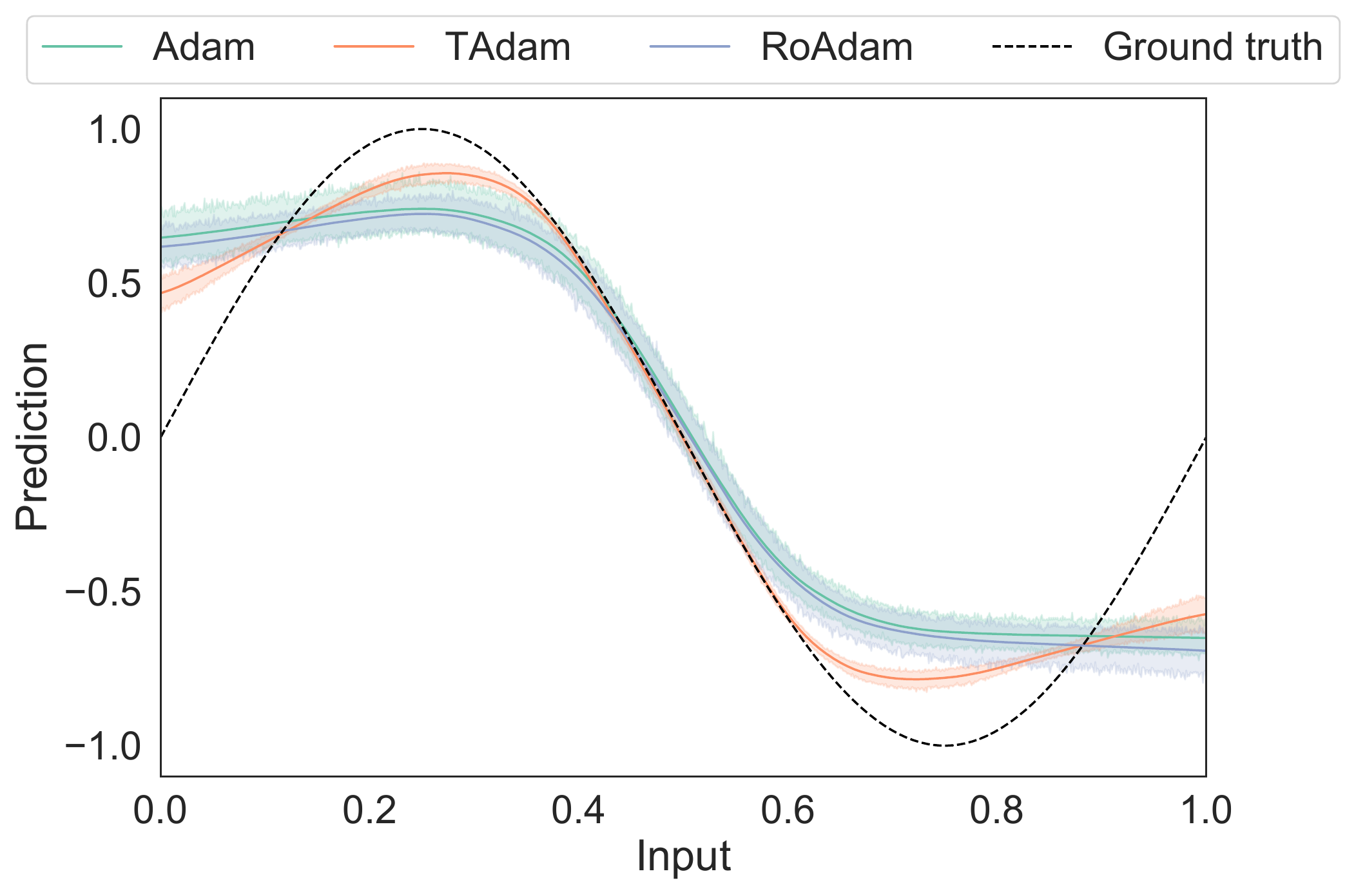}
        \subcaption{$(\nu_{\zeta}, \lambda_{\zeta}) = (1.0, 0.05)$}
        \label{fig:result_lrt_test_1_005}
    \end{minipage}
    \centering
    \begin{minipage}[t]{0.20\linewidth}
        \centering
        \includegraphics[keepaspectratio=true,width=\linewidth]{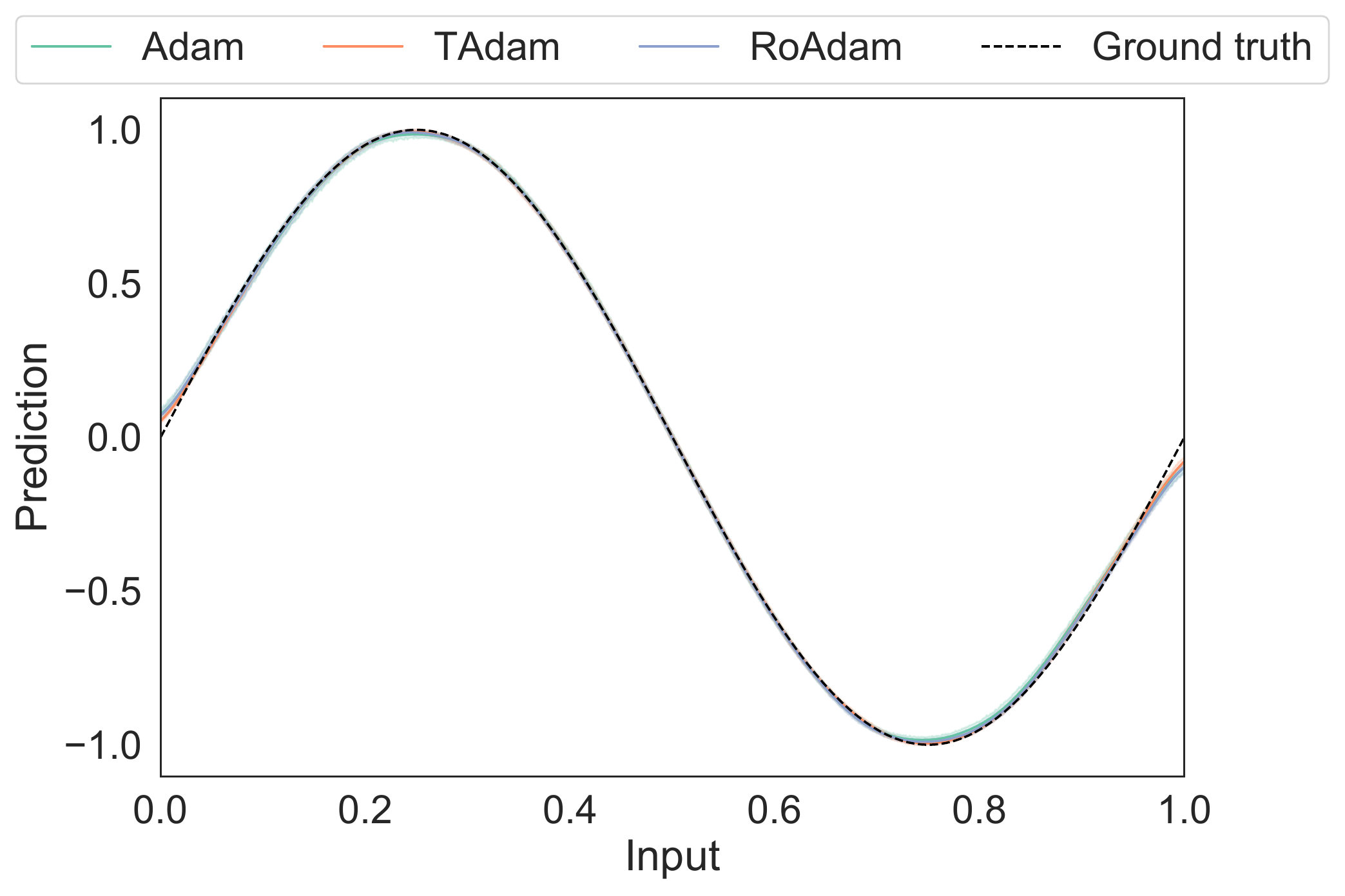}
        \subcaption{$(\nu_{\zeta}, \lambda_{\zeta}) = (2.0, 0.03)$}
        \label{fig:result_lrt_test_2_003}
    \end{minipage}
    \caption{Results of the regression task: (First Two Figures) Loss function w.r.t. the noise probability $p$; in all the noise settings, TAdam outperformed Adam. (Last Two Figures) Prediction curves after learning; although Adam suffered a large variance against the large noise and a bad prediction accuracy, TAdam relatively succeeded in approximating the ground truth function.}
    \label{fig:result_lrt}
\end{figure*}

%Figure classification CIFAR-100
\begin{figure*}[tb]
    \centering
    \begin{minipage}[t]{0.30\linewidth}
        \centering
        \includegraphics[keepaspectratio=true,width=\linewidth]{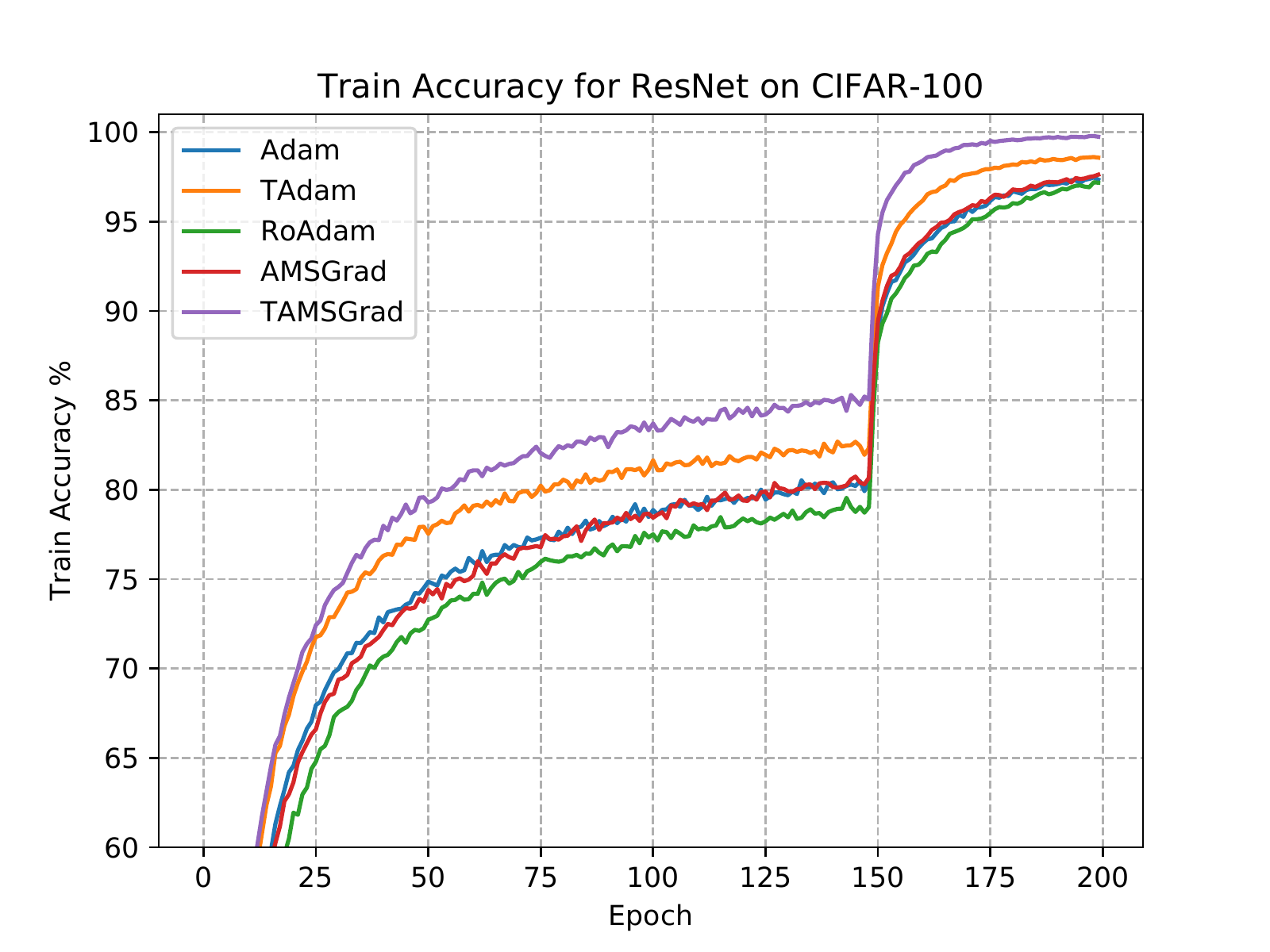}
        \subcaption{Noise-free training accuracy}
        \label{fig:resnet_train}
    \end{minipage}
    \centering
    \begin{minipage}[t]{0.27\linewidth}
        \centering
        \includegraphics[keepaspectratio=true,width=\linewidth]{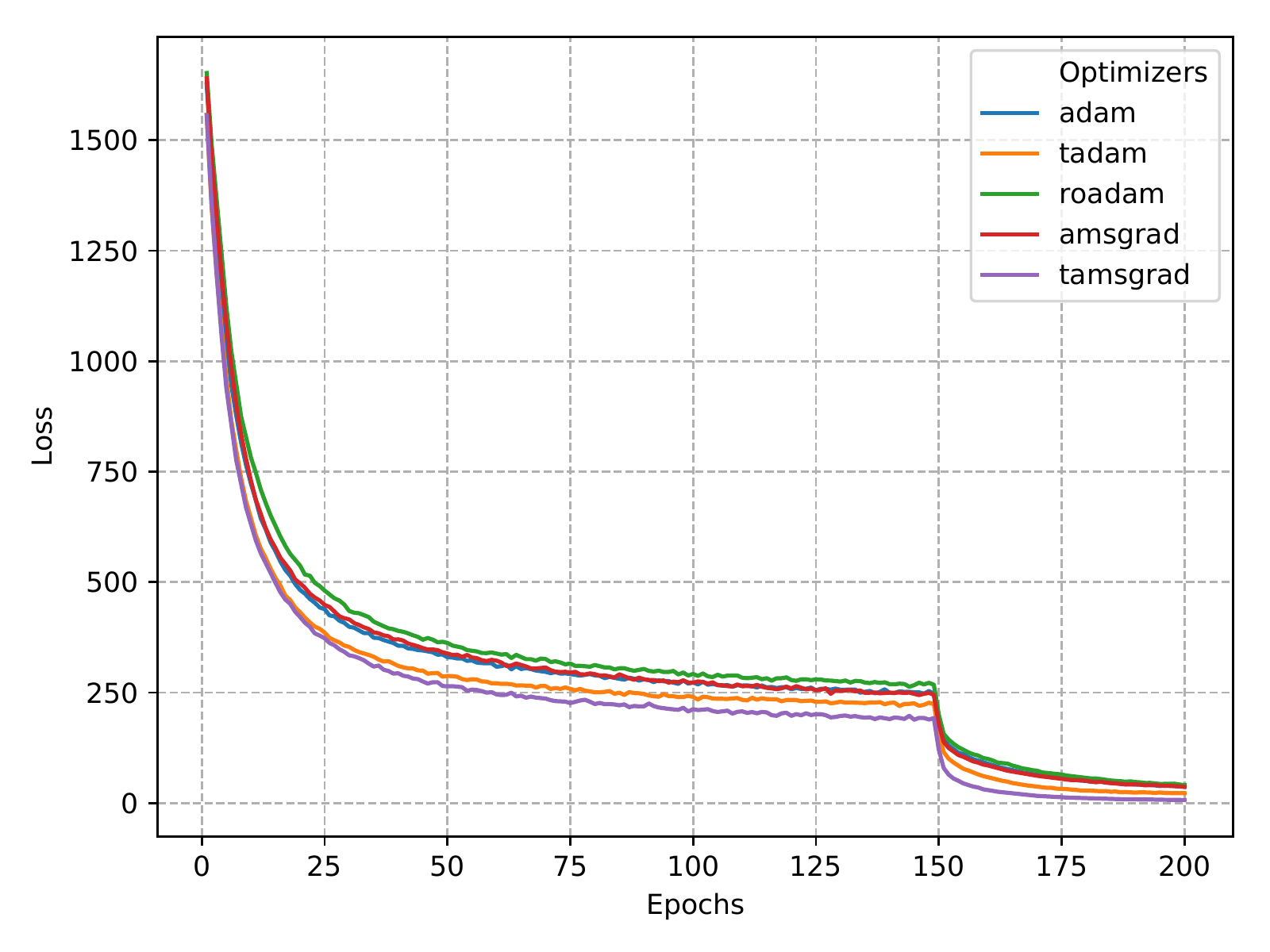}
        \subcaption{Noise-free data training loss per epoch}
        \label{fig:resnet_train_loss}
    \end{minipage}
    \centering
    \begin{minipage}[t]{0.30\linewidth}
        \centering
        \includegraphics[keepaspectratio=true,width=\linewidth]{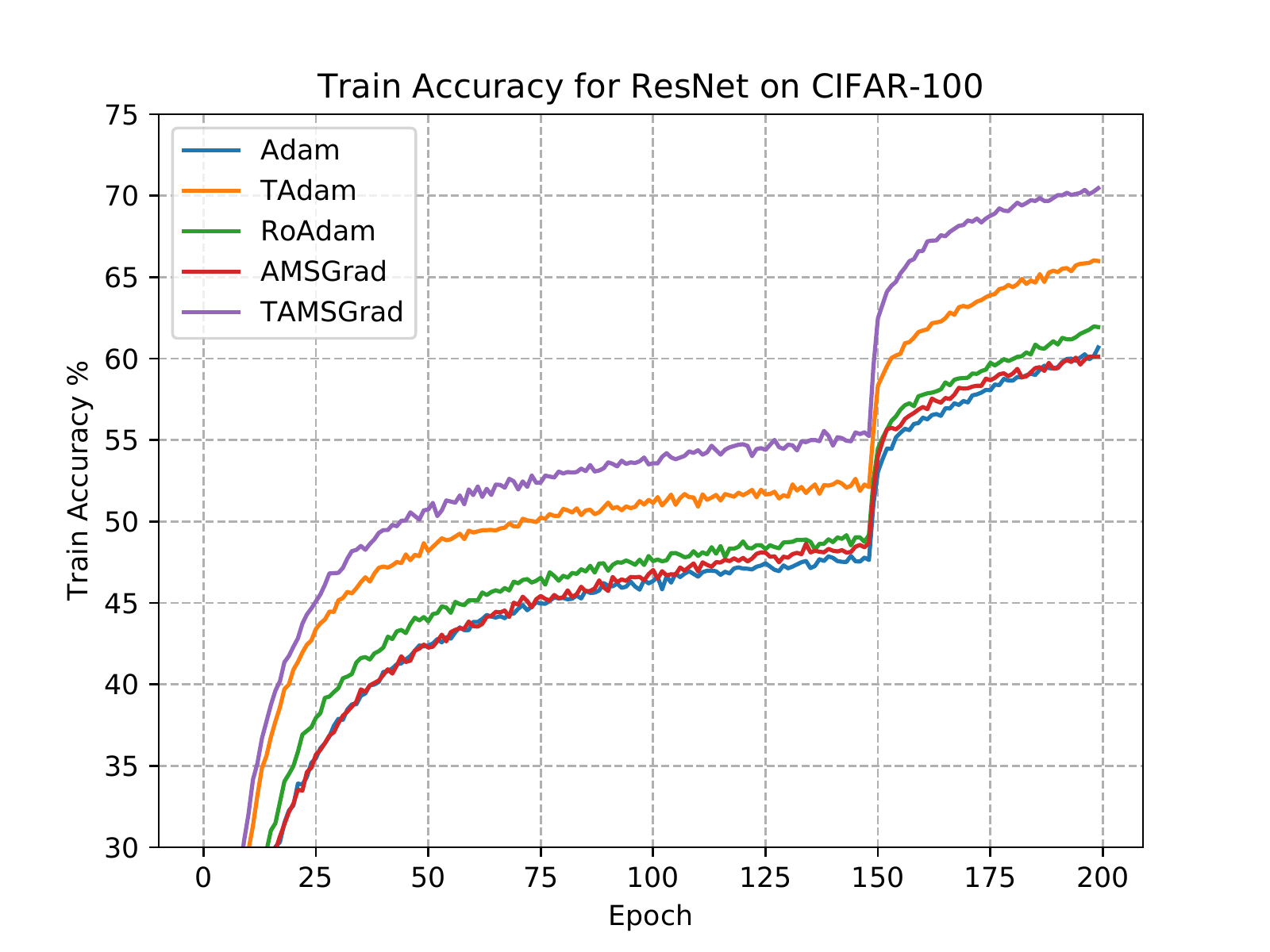}
        \subcaption{Noisy data training accuracy}
        \label{fig:resnet_train_noise}
    \end{minipage}
    \centering
    \begin{minipage}[t]{0.30\linewidth}
        \centering
        \includegraphics[keepaspectratio=true,width=\linewidth]{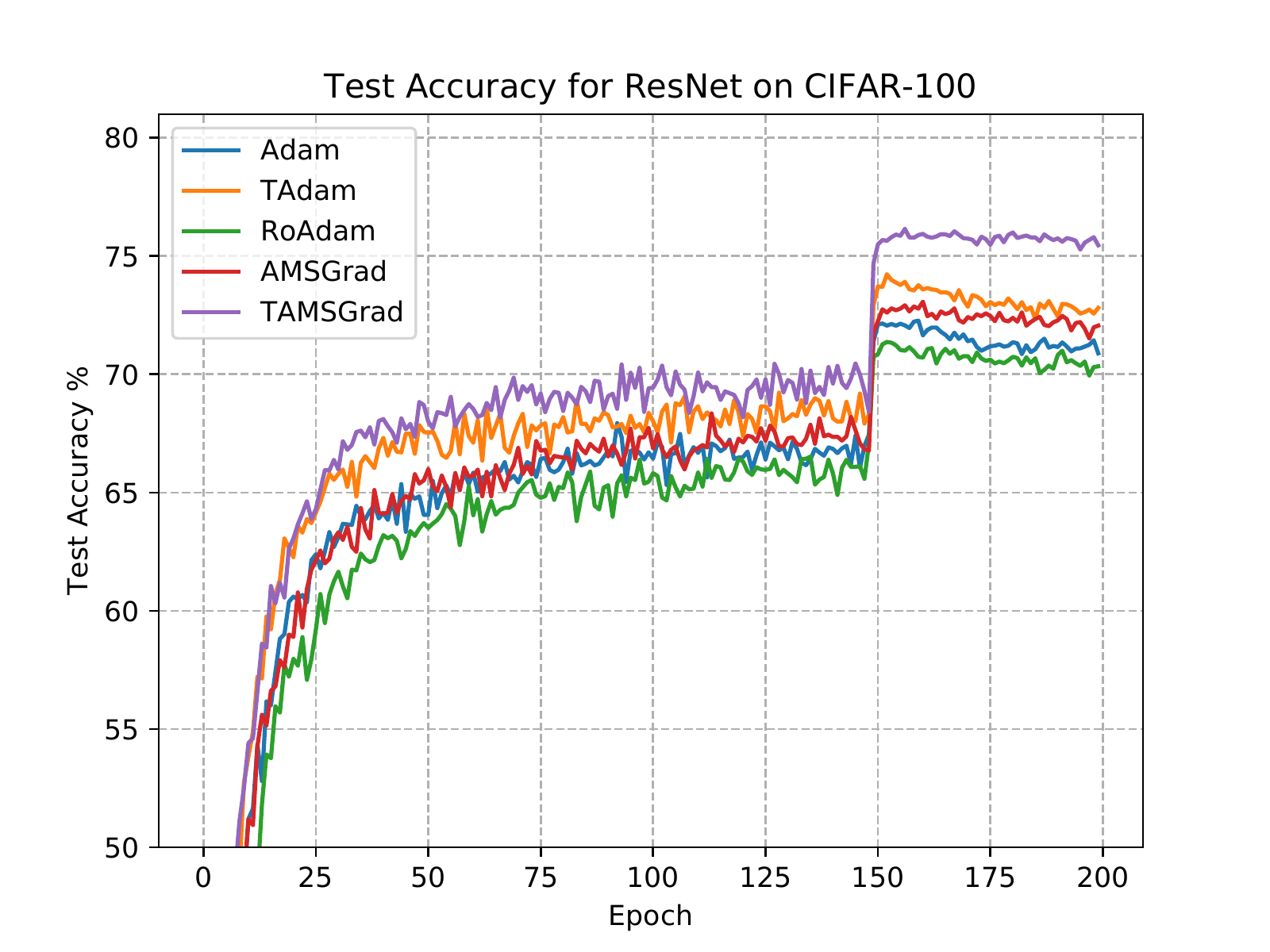}
        \subcaption{Noise-free test accuracy}
        \label{fig:resnet_test}
    \end{minipage}
    \centering
    \begin{minipage}[t]{0.27\linewidth}
        \centering
        \includegraphics[keepaspectratio=true,width=\linewidth]{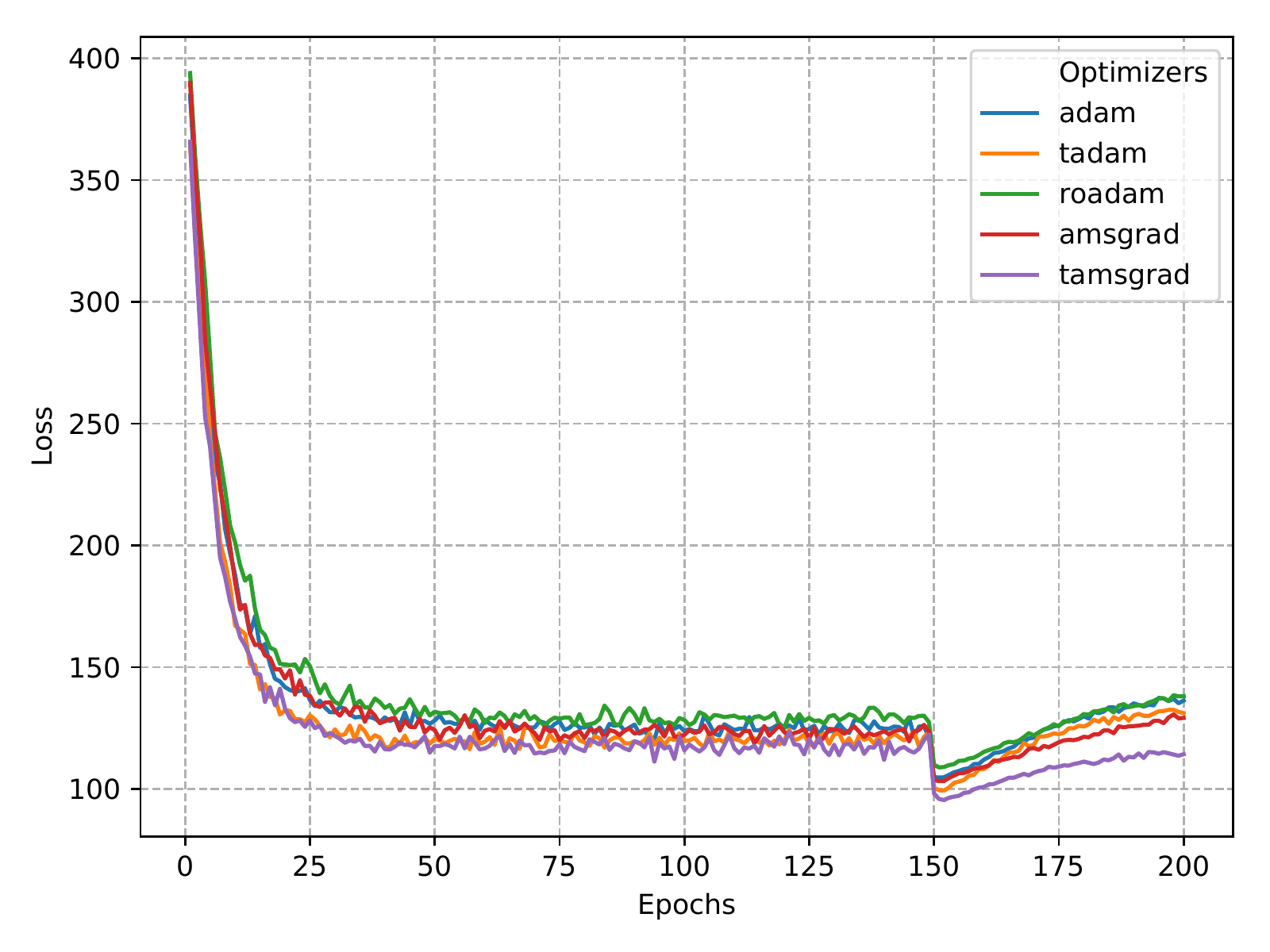}
        \subcaption{Noise-free data test loss per epoch}
        \label{fig:resnet_test_loss}
    \end{minipage}
    \centering
    \begin{minipage}[t]{0.30\linewidth}
        \centering
        \includegraphics[keepaspectratio=true,width=\linewidth]{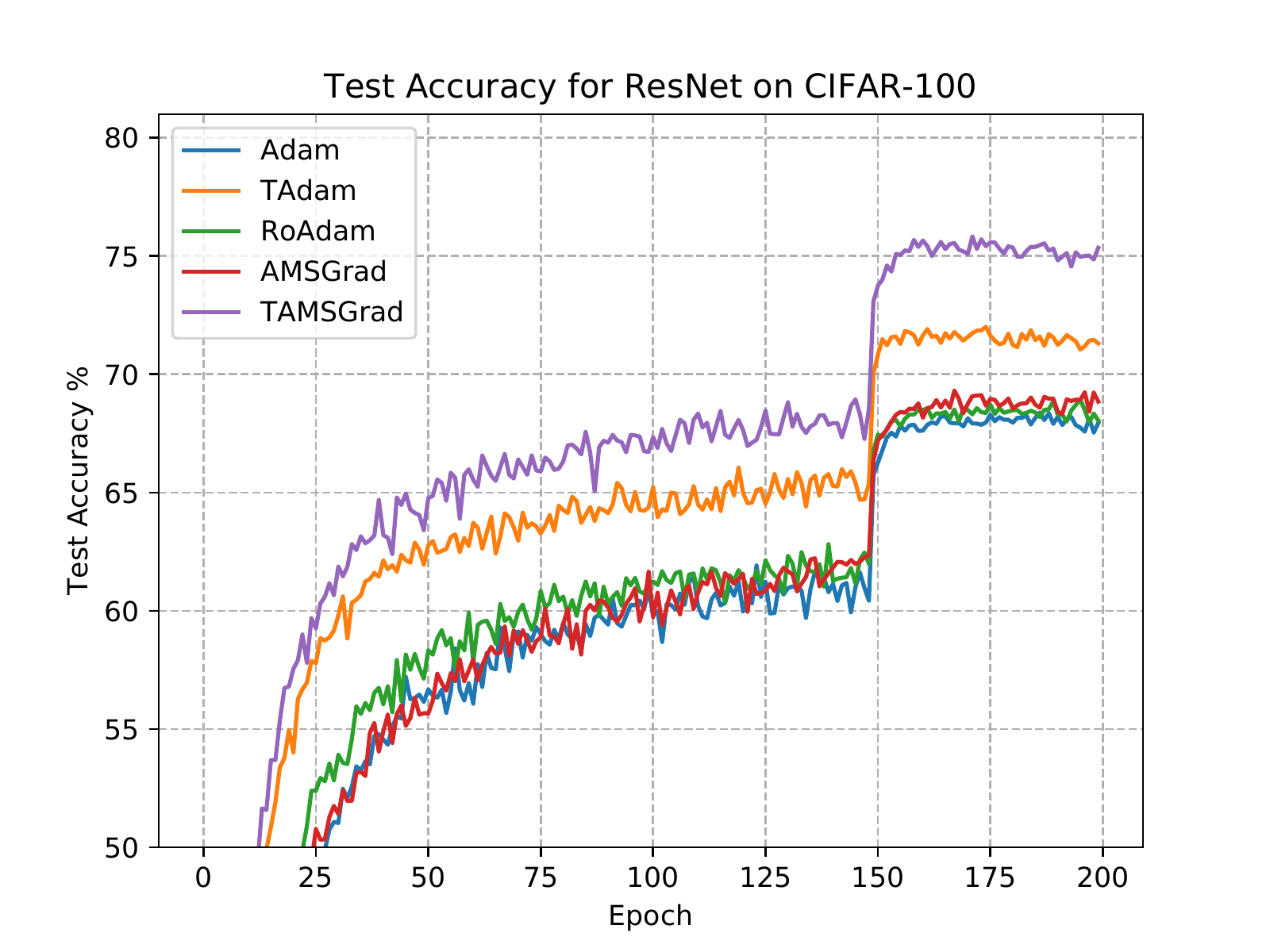}
        \subcaption{Noisy data test accuracy}
        \label{fig:resnet_test_noise}
    \end{minipage}
    \caption{Training and test accuracy (noise-free and noise-included) and loss (noise-free) for ResNet-34 on CIFAR-100.}
    \label{fig:result_cifar100}
\end{figure*}

%Figure Reinforcement Learning
\begin{figure*}[tb]
    \centering
    \begin{minipage}[t]{0.30\linewidth}
        \centering
        \includegraphics[keepaspectratio=true,width=\linewidth]{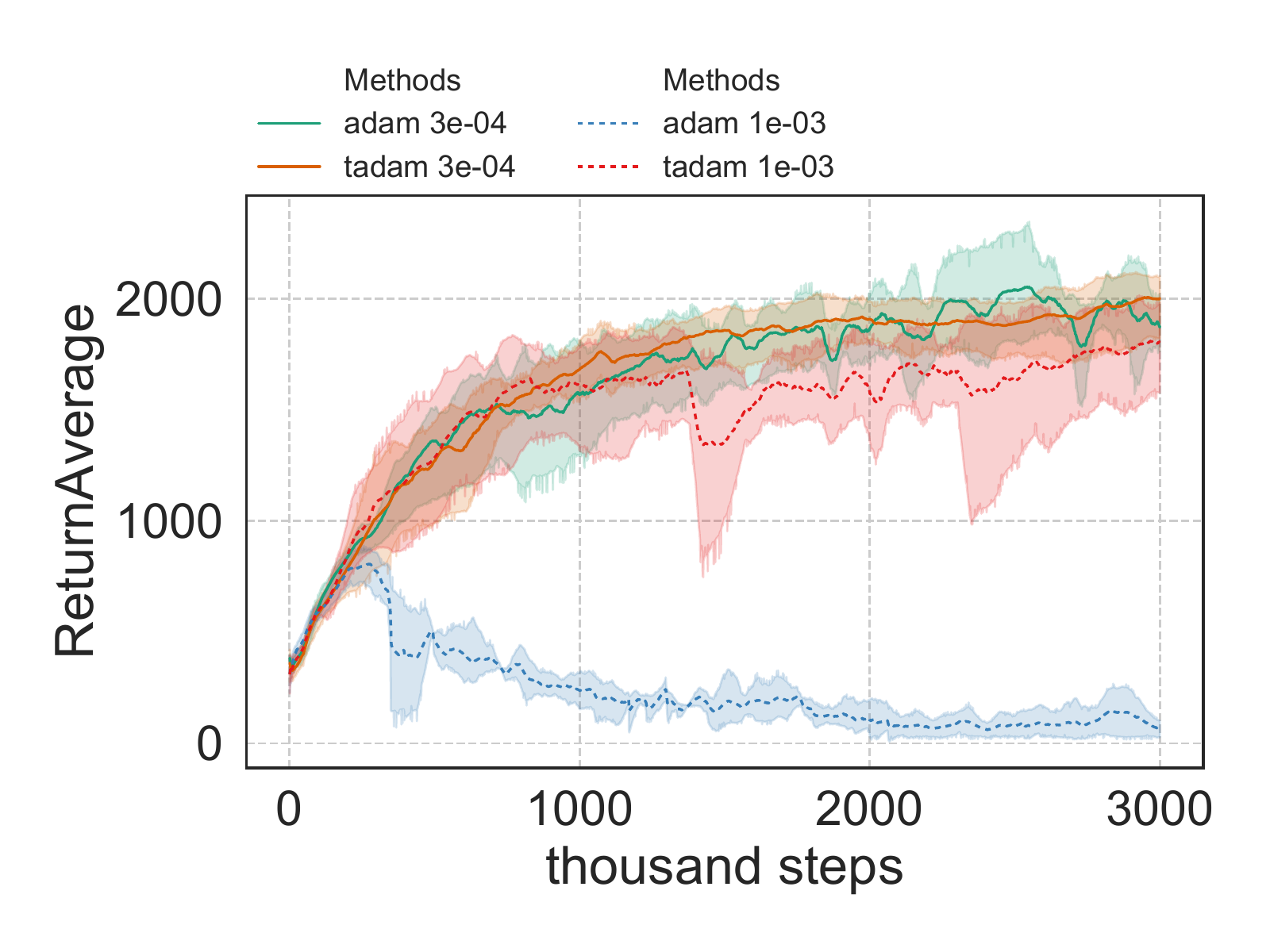}
        \subcaption{Pybullet Ant-v0}
        \label{fig:ant}
    \end{minipage}
    \centering
    \begin{minipage}[t]{0.30\linewidth}
        \centering
        \includegraphics[keepaspectratio=true,width=\linewidth]{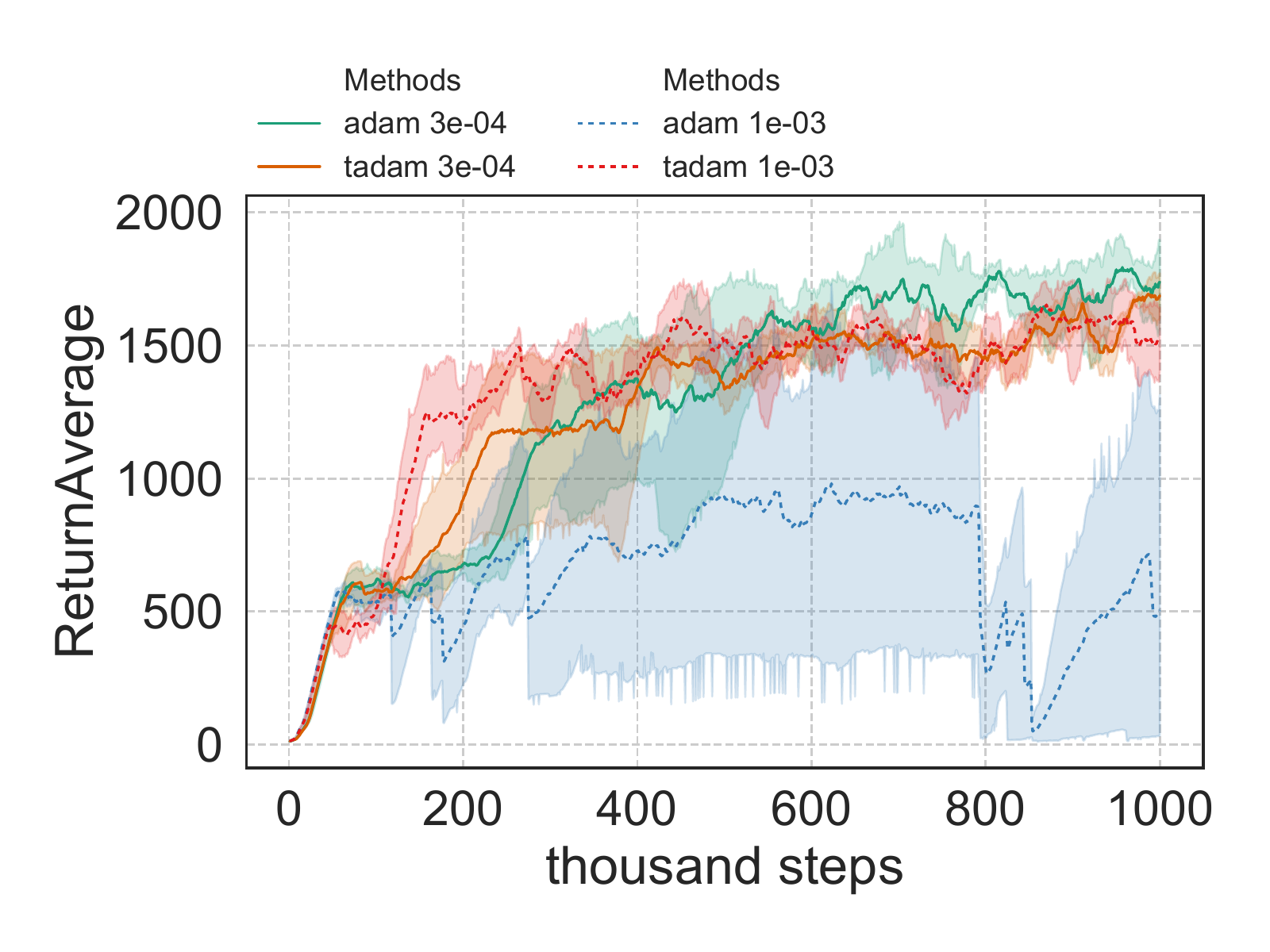}
        \subcaption{Pybullet Hopper-v0}
        \label{fig:hopper}
    \end{minipage}
    \centering
    \begin{minipage}[t]{0.30\linewidth}
        \centering
        \includegraphics[keepaspectratio=true,width=\linewidth]{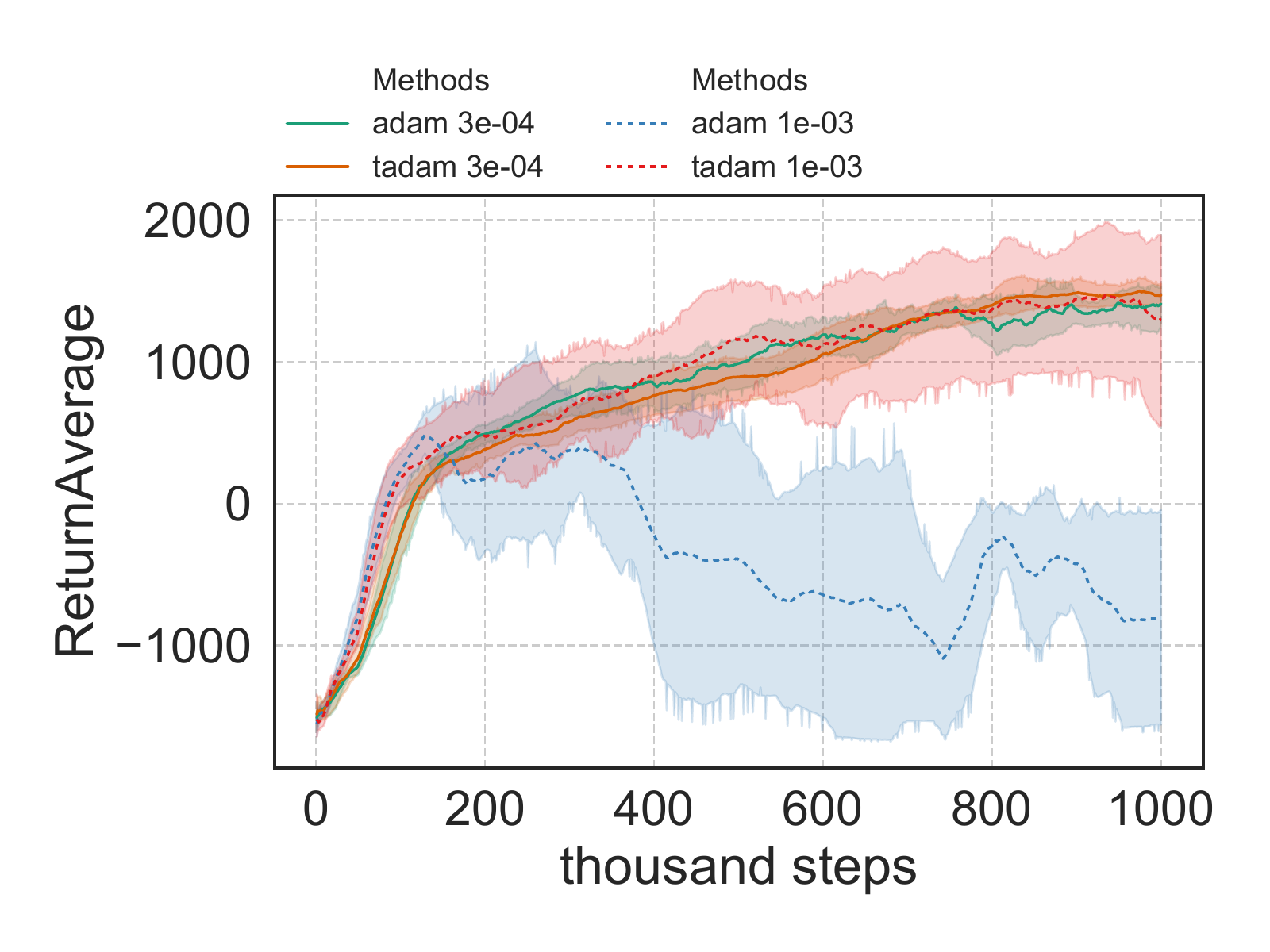}
        \subcaption{Pybullet HalfCheetah-v0}
        \label{fig:halfcheetah}
    \end{minipage}
    \centering
    \begin{minipage}[t]{0.30\linewidth}
        \centering
        \includegraphics[keepaspectratio=true,width=\linewidth]{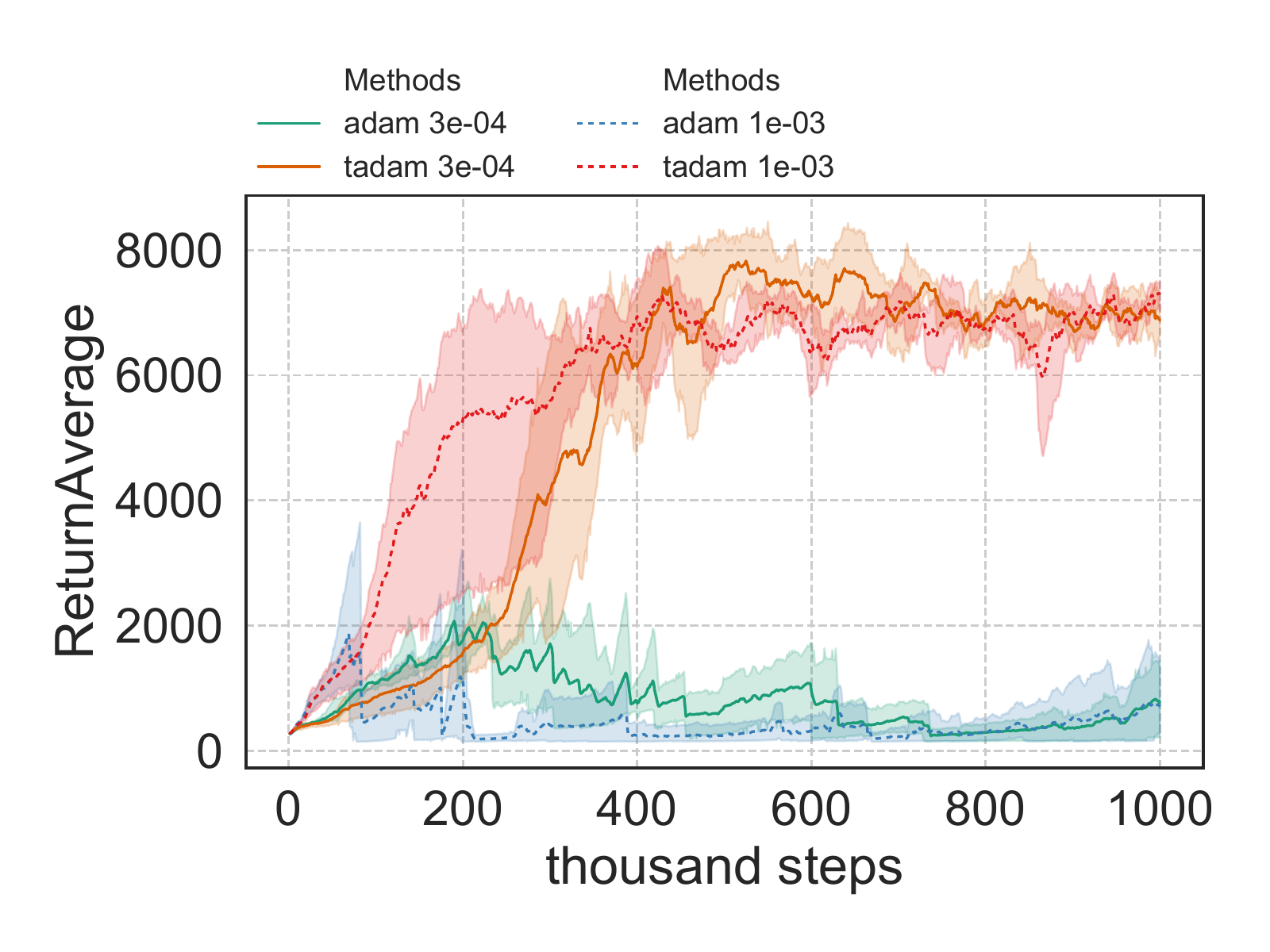}
        \subcaption{Pybullet InvertedDoublePendulum-v0}
        \label{fig:idpendulum}
    \end{minipage}
    \centering
    \begin{minipage}[t]{0.30\linewidth}
        \centering
        \includegraphics[keepaspectratio=true,width=\linewidth]{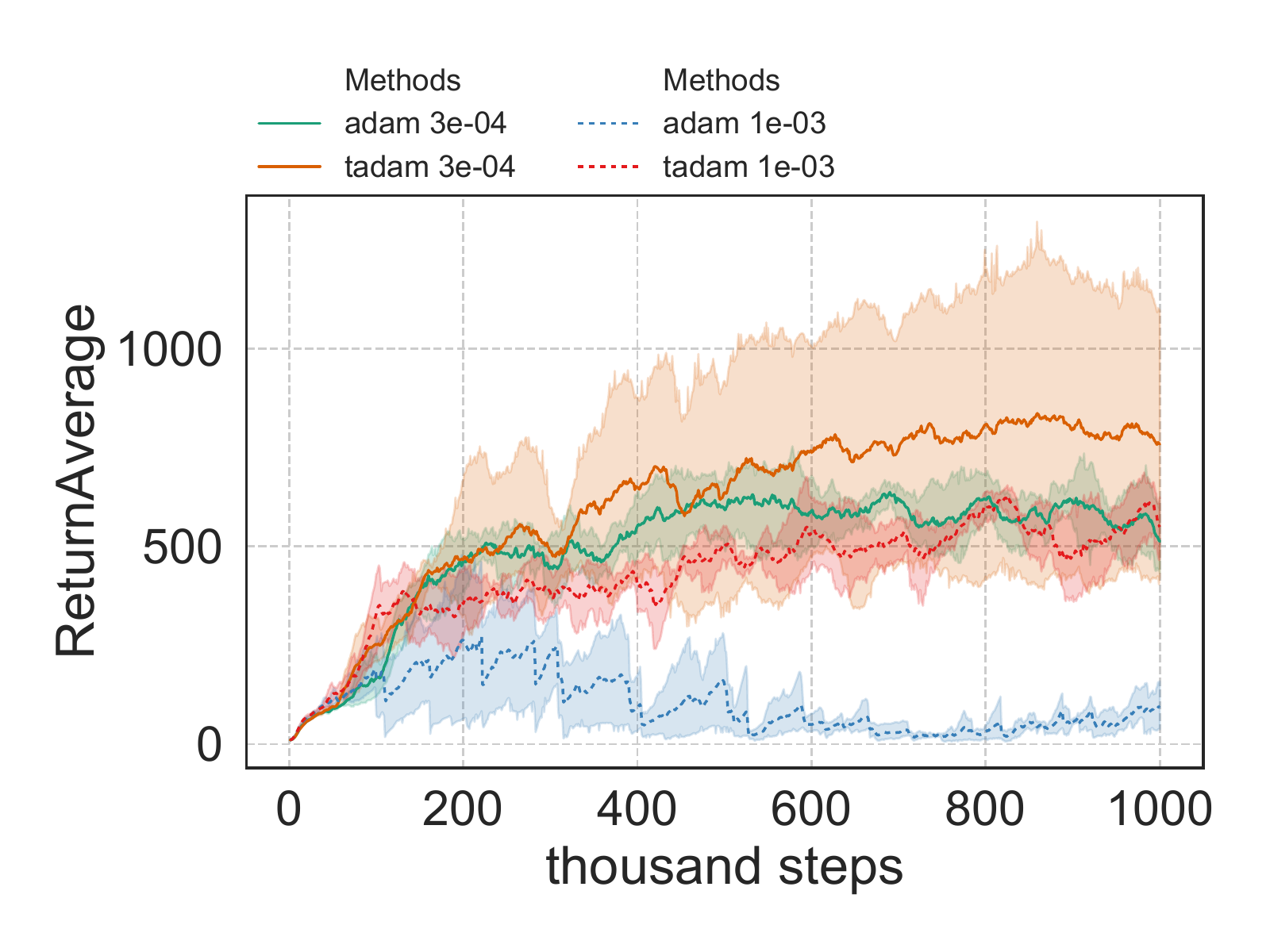}
        \subcaption{Pybullet Walker2D-v0}
        \label{fig:walker2d}
    \end{minipage}
    \centering
    \begin{minipage}[t]{0.30\linewidth}
        \centering
        \includegraphics[keepaspectratio=true,width=\linewidth]{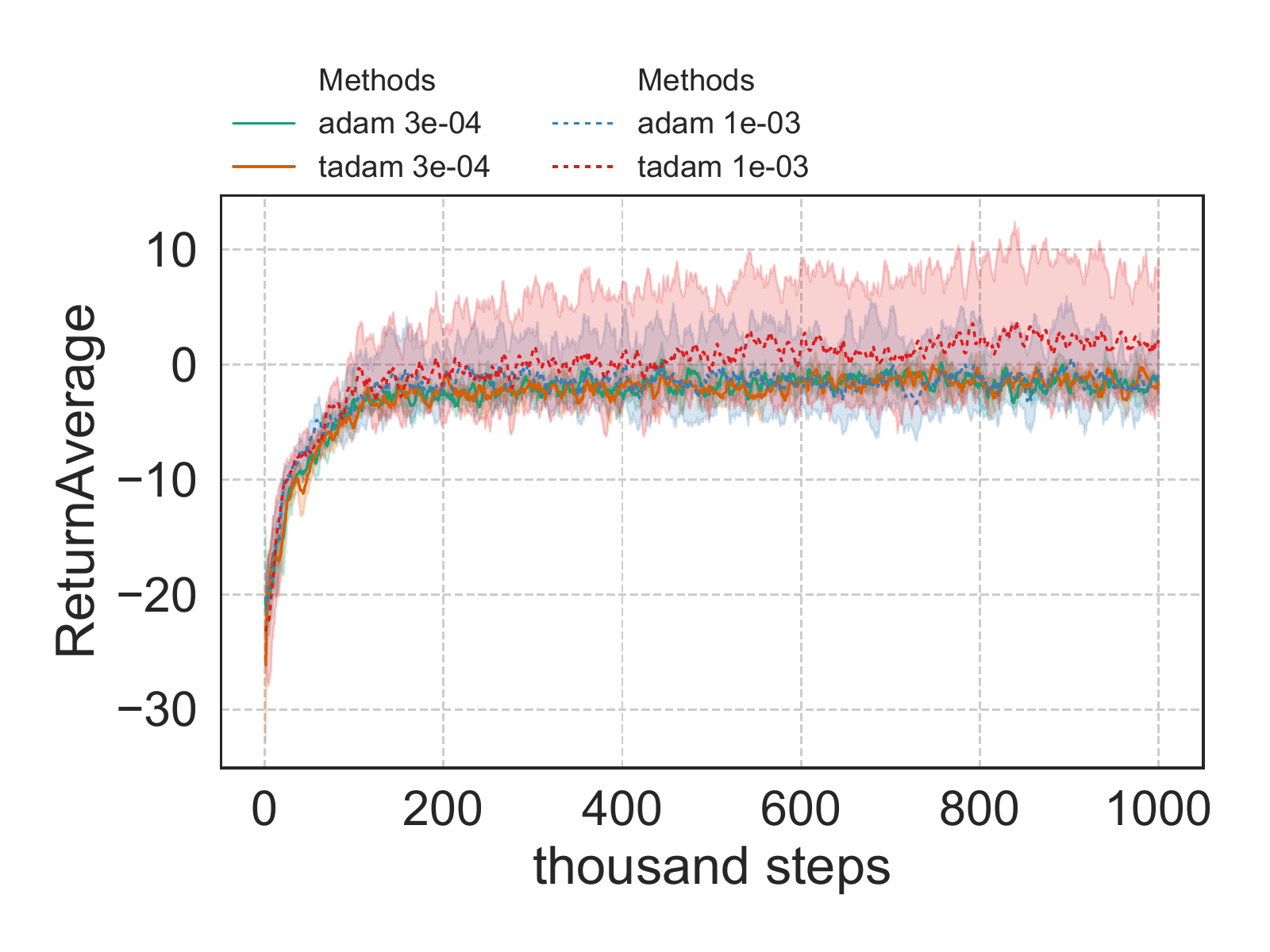}
        \subcaption{Pybullet Reacher-v0}
        \label{fig:reacher}
    \end{minipage}
    \caption{Training curves for PPO agent.}
    \label{fig:rl_ppo_results}
\end{figure*}
 
%%%%%%%%%%%%%%%%%%%%%%%%%%%%%%%%%%%%%%%%
\subsection{Robust Supervised Learning}

 It has been shown~\cite{nettleton2010study} that training standard supervised learning algorithms with noisy data resulted in bad performance and accuracy of the resulting models. In real robotic tasks, it is often unrealistic to assume that the true state is completely observable and noise-free, and perfect supervised signals are difficult to obtain. In the following experiments, TAdam reveals to be useful in increasing the accuracy of the models, even when facing noisy inputs.

%%%%%%%%%%%%%%%%%
\subsubsection{Robust Regression}

\paragraph{Experimental settings}

The regression setting on which we compared TAdam, Adam and RoAdam is as follows.

 A ground truth function is defined as $f(x) = \sin(2\pi x)$ and we set a fully-connected neural network to approximate it from scattered observations $t$, sampled from the true function with noise.
The observations have a probability $p$ of being infected by some noise $\zeta$, so that:
\begin{align}
    t &= \sin(2\pi x) + \zeta\\
    \zeta &\sim \mathrm{St}(\nu_{\zeta},  0, \lambda_{\zeta}) \mathrm{Bern} \left ( \frac{p}{100} \right )\\
    p &\in \{0, 10, 20, 30, 40, 50, 60, 70, 80, 90, 100\}
\end{align}
where $\mathrm{St}(\nu_{\zeta}, \lambda_{\zeta})$ designates a student-t distribution with degrees of freedom $\nu_{\zeta}$, $0$ location, and scale $\lambda_{\zeta}$, and $\mathrm{Bern}(p/100)$ is a Bernoulli distribution with the probability $p$ as its parameter.

The model, on the other hand, is a neural network with $5$ linear layers, each composed of $50$ neurons.
The ReLU activation function~\cite{relu} is used for all the hidden layers, while the loss function for the network is the Mean Squared Error (MSE).

\paragraph{Experimental results}

The results of the loss functions against the noise probability $p$ on the regression task are depicted in Fig.~\ref{fig:result_lrt_1_005} and Fig.~\ref{fig:result_lrt_2_003}.
Note that 50 trials are conducted for each $p$.
As it can be seen, TAdam absolutely outperformed Adam in all the cases, and reveals to be more robust than RoAdam.
 In addition, as the noise probability in the observations increases, TAdam managed to resist to their effect.

To visualize the learning results, the predicted curves after learning are also illustrated in Fig.~\ref{fig:result_lrt_test_1_005} and Fig.~\ref{fig:result_lrt_test_2_003}.
The learning variances of Adam were obviously larger than those generated by TAdam, and TAdam relatively succeeded in following the ground truth function from the observations even with large noise.

%%%%%%%%%%%%%%%%%
\subsubsection{Robust Classification}

\paragraph{Experimental settings}

 Here, we use the same experimental settings described in~\cite{adabound} and compare Adam, AMSGrad and their T versions along with RoAdam on an image classification task on the standard CIFAR-100 dataset.
 
 The architecture of the convolutional network involved in the described experiments is the ResNet-34~\cite{resnet34}. A fixed budget of 200 epochs are used throughout the training, and the learning rates are reduced by 10 after 150 epochs.
 
 The optimizers are launched with the following hyperparameter values: \{learning rate: 0.001\}, \{betas: (0.99, 0.999)\} and both T algorithms use the default degrees of freedom, i.e. \{degrees of freedom = dimension of the gradients\}. The third beta value of RoAdam is also set to \{0.999\}.

\paragraph{Experimental results}
  
  We first launched a simulation without noise, using directly the unmodified datasets. The results for that simulation are found in Fig.~\ref{fig:resnet_train} and Fig.~\ref{fig:resnet_test}. We can see that TAdam and TAMSGrad are able to achieve faster convergence during the training phase compared to the standard versions, and also show higher level of generalization during the test phase. The corresponding loss curves, Fig.~\ref{fig:resnet_train_loss} and Fig.~\ref{fig:resnet_test_loss}, show that TAMSGrad is able to reach a lowest point during the training phase, while also keeping a low loss value on the test data. 
 This result points the fact that TAMSGrad builds on the combined improvement of the first moment (TAdam) and second moment (AMSGrad) in order to provide a more stable algorithm that can outperform the others.
  
 Next, we applied, with a probability of 25\%, a color jittering effect on the training dataset and replaced 20\% of the original training data points by fake ones, in order to test the ability of the optimizers to extract the most useful informations from corrupted datasets. The results can be seen in Fig.~\ref{fig:resnet_train_noise} and Fig.~\ref{fig:resnet_test_noise} and it highlights the benefits of TAdam against Adam. Indeed, even thought the value of beta1 is larger (0.99 instead of default 0.9), Adam remains sensitive to outliers, while TAdam can ignore them.

%%%%%%%%%%%%%%%%%%%%%%%%%%%%%%%%%%%%%%%%
\subsection{Robust Reinforcement Learning}

 Whether it comes from sensors, or from bad estimates during learning, or from different feedbacks from different human instructors (e.g. non-technical users in real-world robotics situations), noisiness is inseparable from robotics reinforcement learning (RL). In order to test the robustness properties of TAdam in RL tasks, we conducted some simulations on six different Pybullet gym environments~\cite{coumans2019}. The results are summarized in Fig.~\ref{fig:rl_ppo_results}.

\paragraph{Experimental settings}

 The Fig.~\ref{fig:rl_ppo_results} summarizes four trials with four different seeds on each environment. The algorithm employed is the proximal policy optimization (PPO)~\cite{ppo}, from the Berkley artificial intelligence research implementation, rlpyt~\cite{stooke2019rlpyt}, with the following setting:

\begin{table}[h]
\begin{center}
\begin{tabular}{|c||c|}
\hline
Value loss coef. & 1\\
\hline
Entropy loss coef. & 0\\
\hline
GAE parameter $\lambda$  & 0.95\\
\hline
Num. Epochs  & 10\\
\hline
Ratio clipping $\epsilon$  & 0.2\\
\hline
Horizon T  & 2048\\
\hline
Minibatch size  & 64\\
\hline
TAdam d.o.f. $\nu$  & $dim(g)$\\
\hline
\end{tabular}
\end{center}
\caption{Settings for the RL experiments}
\label{table_rlsetting}
\end{table}

No gradient norm clipping was used throughout the simulations, since the property at test is the robustness of the optimizers to aberrant gradient values and their ability to produce good policies. Gradient norm clipping introduces a manually defined heuristic threshold, which depends on the task and on various conditions, and moreover, is used for the norm of all gradients larger than its value. Such trick would therefore introduce some undesirable bias in the results.

 The simulations involved two different learning rates: the widely used and fine tuned value for Adam, $3 \times 10^{-4}$, and the defined default, yet larger value, $1 \times 10^{-3}$.

\paragraph{Experimental results}

 Searching for the optimal learning rate is commonly known to be a tedious and serious problem in SGD based algorithms, and high learning rates (particularly the default Adam step value $1 \times 10^{-3}$) are usually not used in reinforcement learning due to the amount of noise coming from the early bootstrapping stage, but also to avoid the agent from reaching an early deterministic policy.
 
 As displayed by the results in Fig.~\ref{fig:rl_ppo_results}, a high learning rate causes Adam to suffer from both these problems and makes it unable to converge to a good policy. On the other hand, TAdam proves to be robust enough to sustain different learning rates, and learns the tasks with both given hyperparameter values. Thanks to its careful updates of the agent, TAdam can still reach a sub-optimal policy that may even be better than the one reached with smaller learning rates (Fig.~\ref{fig:halfcheetah}, \ref{fig:reacher}). This feature offered by TAdam not only allows for the use of higher learning rates in order to accelerate the learning process, but also reduces the difficulties related to the tuning of the learning rate, since the default learning rate can be directly used.

 Also, as stated in the experimental settings section, no gradient norm clipping was used during the simulations. Without this trick, we can see that Adam fails altogether on the inverted double pendulum task, while TAdam naturally and automatically ignores or reduces the effect of large gradients, keeping the gradient (momentum) from overshooting during learning and making the gradient norm clipping stratagem unnecessary.

%%%%%%%%%%%%%%%%%%%%%%%%%%%%%%%%%%%%%%%%%%%%%%%%%%%%%%%%%%%%%%%%%%%%%%%%%%%%%%%%
\section{Conclusion and Future Work}

 In this letter, we proposed and described TAdam, a new stochastic gradient optimizer, which makes the Adam algorithm much more robust and provides a way to produce stable and efficient machine learning applications.
 TAdam is based on the robust mean estimate rule of the Student-t distribution as an alternative to the standard EMA.
 We verified that TAdam outperformed Adam in terms of robustness on supervised learning (regression and classification) tasks, and reinforcement learning tasks.

 In this work, TAdam uses a fixed degrees of freedom $\nu$ which is equal to the dimension of the gradients, and therefore has a fixed robustness. A straightforward improvement is therefore to design a mechanism that automatically updates the parameter $\nu$ during the learning process, according to the presence or absence of outliers.

%%%%%%%%%%%%%%%%%%%%%%%%%%%%%%%%%%%%%%%%%%%%%%%%%%%%%%%%%%%%%%%%%%%%%%%%%%%%%%%%

%%%%%%%%%%%%%%%%%%%%%%%%%%%%%%%%%%%%%%%%%%%%%%%%%%%%%%%%%%%%%%%%%%%%%%%%%%%%%%%%
\section*{APPENDIX}

\subsection{Proof of Theorem 1}
First, we start by noticing that the basic bound of the regret from the convergence proof by Reddi et al.~\cite{reddi2019amsgrad}  also holds for TAdam, i.e.:
\begin{align}
  \begin{split}
 &R_T = \sum\limits^T_{t=1} f_t(\theta_t) - f_t(\theta*) \leq \\
 &\sum\limits^T_{t=1}  \left.\lbrace\frac{\left[ \parallel \hat{V}_t^{1/4}(\theta_t - \theta*)\parallel^2 - \parallel \hat{V}_t^{1/4}(\theta_{t+1} - \theta*)\parallel^2 \right]}{2\alpha_t (1-\beta_{1t})} \right.\\
  &\left.  + \frac{\alpha_t \parallel \hat{V}_t^{-1/4} m_t \parallel^2}{(1-\beta_{1t})} + \frac{\beta_{1t} \parallel \hat{V}_t^{1/4}(\theta_t - \theta*)\parallel^2}{2\alpha_t (1-\beta_{1t})} \rbrace \right.
  \end{split}
  \label{eq:init_regret_bound}
 \end{align}

 However, to further refine this upper bound, we need to redefine the \textit{Lemma 2} used in the proof of Reddi et al, since $\beta_{1t} = \beta_w = \frac{W_{t-1}}{W_{t-1} + w_t} \leq \beta_1$ does not hold anymore for all time step $t$. For this purpose, we use the expected value of $\beta_w$, $\bar{\beta}_w = \mathbb{E}[\beta_w] < 1$, instead of $\beta_1$, to define the upper bound and, following the same process as Reddi et al., define a similar expression to their \textit{Lemma 2} in the case of TAdam:
\begin{align}
&\sum\limits^T_{t=1} \alpha_t \parallel \hat{V}_t^{-1/4} m_t \parallel^2 \nonumber \\
&\leq \frac{\alpha \sqrt{1 + \log T}}{(1-\bar{\beta}_w)(1 - \gamma)\sqrt{(1-\beta_2)}} \sum\limits^d_{i=1} \parallel g_{1:T, i}\parallel_2
\label{eq:lemma2}
\end{align}

 Based on this new lemma, the remaining steps are completely identical to the proof of Reddi et al., and the final regret bound of TAdam is given by:
 \begin{align}
 \begin{split}
 R_T &\leq \left.\frac{D^2_{\infty}}{2\alpha_T (1-\bar{\beta}_w)} \sum\limits_{i=1}^d \hat{v}_{T,i}^{1/2} + \frac{D^2_{\infty}}{(1-\bar{\beta}_w)^2} \sum\limits^T_{t=1} \sum\limits_{i=1}^d \frac{\beta_{1t}\hat{v}_{t,i}^{1/2}}{\alpha_t} \right.\\
&\left. + \frac{\alpha \sqrt{1 + \log T}}{(1-\bar{\beta}_w)^2(1 - \gamma)\sqrt{(1-\beta_2)}} \sum\limits^d_{i=1} \parallel g_{1:T, i}\parallel_2 \right.
  \end{split}
  \label{eq:regret_bound}
 \end{align}

\subsection{Proof of Theorem 2}
 Assuming that the gradients $g$ ultimately follow an asymptotic normal distribution $g \in \mathbb{R}^d \sim \mathcal{N}(\mu, \Sigma)$, then we know that $D_M^2(g, \mu) = \sum\limits_j \frac{(g_t^j - \mu_j)^2}{v_j} \sim \chi^2(d)$, where $d$ is the degrees of freedom of the chi-squared distribution. Applying this to the Mahalanobis distance in TAdam, we have:
  \begin{align}
  D_t &= \sum\limits_j \frac{(g_t^j - m_{t-1}^j)^2}{v_{t-1}} \sim \chi^2(d)
  \end{align}
  
  Now, we know that the expected value of the chi-squared distribution with $d$ degrees of freedom is $\mathbb{E}[D_t] = d$ and the expected value of the inverse-chi-squared distribution with the same degrees of freedom is given by $\mathbb{E}[D^{-1}_t] = \frac{1}{d - 2},  \forall d > 2$. We can therefore define:
 \begin{align*}
  E_{min} &\leq \mathbb{E}\left[\frac{1}{D_t + \nu}\right] \leq E_{max}\\
  E_{max} &= \frac{1}{\mathbb{E}[D^{-1}_t]^{-1} + \nu} = \frac{1}{d - 2 + \nu} \leq \frac{1}{d - 2}\\
  E_{min} &= \frac{1}{\mathbb{E}[D_t] + \nu} = \frac{1}{d + \nu}
 \end{align*}

 This inequality comes from the Jensen's inequality and from the fact that $f(x) = \frac{1}{x + \nu}$ and $f(x) = \frac{1}{x^{-1} + \nu}$ are respectively convex and concave. The expected value of the weights $w_t$ in TAdam, can therefore be expressed as:
 \begin{align}
 1 \leq \mathbb{E}[w_t] = \mathbb{E} \left [\frac{\nu + d}{\nu + D_t} \right ] \leq \frac{\nu + d}{d - 2}
 \end{align}
 We can then infer the mean of the weighted sum $W_t$:
 \begin{align}
 W_{t} &= (\frac{\beta_1}{1-\beta_1})[(\frac{2\beta_1 - 1}{\beta_1})^{t-1}] + \sum\limits_{i=1}^{t-1} w_i (\frac{2\beta_1 - 1}{\beta_1})^{t-1-i} \nonumber \\
 \mathbb{E}[W_t] &= (\frac{\beta_1}{1-\beta_1})[a^{t-1}] + (\frac{\beta_1}{1-\beta_1}) \mathbb{E}[w_t] (1 - a^{t-1}) \nonumber \\
  &\leq (\frac{\beta_1}{1-\beta_1})\left(a^{t-1} + \mathbb{E}[w_t] (1 - a^{t-1})\right) \nonumber \\
  &\leq (\frac{\beta_1}{1-\beta_1})\mathbb{E}[w_t] \leq (\frac{\beta_1}{1-\beta_1})\left(\frac{\nu + d}{d - 2}\right) \label{eq:relationWtow}
 \end{align}
 Where we have defined $a = \frac{2\beta_1 - 1}{\beta_1}$ and taken advantage of the monotonic decrease of the sequence $a^t$ towards $0$, given that $a < 1$ for $\beta_1 < 1$. We move on to express the upper bound for $\mathbb{E}[\beta_w]$ where $\beta_w = \frac{W_{t-1}}{W_{t-1} + w_t}$. For this purpose, we make use of the Hartley and Ross unbiased estimator for the mean of the ratio between two random variables~\cite{hartley1954unbiased, goodman1958precision}, which, based on the fact that the covariance between $W_{t-1}$ and $W_{t-1} + w_t$ is positive, gives:
\begin{align}
  \mathbb{E}[\beta_w] &\leq \frac{\mathbb{E}[W_{t-1}]}{\mathbb{E}[W_{t-1} + w_t]} \nonumber \\
 &\leq \frac{(\frac{\beta_1}{1-\beta_1})\left(\frac{\nu + d}{d - 2}\right)}{(\frac{\beta_1}{1-\beta_1})\left(\frac{\nu + d}{d - 2}\right) + (\frac{\nu + d}{d - 2})} = \beta_1
 \label{eq:bw_avg_bound}
  \end{align}
  The last inequality is drawn from the relations depicted by Eq.~\ref{eq:relationWtow}.
 
\addtolength{\textheight}{0cm}%{-12cm}   % This command serves to balance the column lengths

%%%%%%%%%%%%%%%%%%%%%%%%%%%%%%%%%%%%%%%%%%%%%%%%%%%%%%%%%%%%%%%%%%%%%%%%%%%%%%%%

%%%%%%%%%%%%%%%%%%%%%%%%%%%%%%%%%%%%%%%%%%%%%%%%%%%%%%%%%%%%%%%%%%%%%%%%%%%%%%%%
\bibliographystyle{ieeetr}
\bibliography{bibliography}

\begin{thebibliography}{10}

\bibitem{bottou2010large}
L.~Bottou, ``Large-scale machine learning with stochastic gradient descent,''
  in {\em Proceedings of COMPSTAT'2010}, pp.~177--186, Springer, 2010.

\bibitem{sgd_paper}
H.~Robbins and S.~Monro, ``A stochastic approximation method,'' {\em The annals
  of mathematical statistics}, pp.~400--407, 1951.

\bibitem{deep_learning}
Y.~LeCun, Y.~Bengio, and G.~Hinton, ``Deep learning,'' {\em nature}, vol.~521,
  no.~7553, p.~436, 2015.

\bibitem{adam}
D.~P. Kingma and J.~Ba, ``Adam: A method for stochastic optimization,'' {\em
  arXiv preprint arXiv:1412.6980}, 2014.

\bibitem{suchi2019easylabel}
M.~Suchi, T.~Patten, D.~Fischinger, and M.~Vincze, ``Easylabel: A
  semi-automatic pixel-wise object annotation tool for creating robotic rgb-d
  datasets,'' in {\em International Conference on Robotics and Automation
  (ICRA)}, pp.~6678--6684, IEEE, 2019.

\bibitem{gupta2018robot}
A.~Gupta, A.~Murali, D.~P. Gandhi, and L.~Pinto, ``Robot learning in homes:
  Improving generalization and reducing dataset bias,'' in {\em Advances in
  Neural Information Processing Systems}, pp.~9094--9104, 2018.

\bibitem{prml}
C.~M. Bishop, {\em Pattern recognition and machine learning}.
\newblock springer, 2006.

\bibitem{sgd_momentum}
B.~T. Polyak, ``Some methods of speeding up the convergence of iteration
  methods,'' {\em USSR Computational Mathematics and Mathematical Physics},
  vol.~4, no.~5, pp.~1--17, 1964.

\bibitem{nag_paper}
Y.~Nesterov, ``A method for unconstrained convex minimization problem with the
  rate of convergence o (1/k\^{} 2),'' in {\em Doklady AN USSR}, vol.~269,
  pp.~543--547, 1983.

\bibitem{sag_paper}
N.~L. Roux, M.~Schmidt, and F.~R. Bach, ``A stochastic gradient method with an
  exponential convergence \_rate for finite training sets,'' in {\em Advances
  in neural information processing systems}, pp.~2663--2671, 2012.

\bibitem{svrg_paper}
R.~Johnson and T.~Zhang, ``Accelerating stochastic gradient descent using
  predictive variance reduction,'' in {\em Advances in neural information
  processing systems}, pp.~315--323, 2013.

\bibitem{adagrad}
J.~Duchi, E.~Hazan, and Y.~Singer, ``Adaptive subgradient methods for online
  learning and stochastic optimization,'' {\em Journal of Machine Learning
  Research}, vol.~12, no.~Jul, pp.~2121--2159, 2011.

\bibitem{adadelta}
M.~D. Zeiler, ``Adadelta: an adaptive learning rate method,'' {\em arXiv
  preprint arXiv:1212.5701}, 2012.

\bibitem{rmsprop}
T.~Tieleman and G.~Hinton, ``Lecture 6.5-rmsprop: Divide the gradient by a
  running average of its recent magnitude,'' {\em COURSERA: Neural networks for
  machine learning}, vol.~4, no.~2, pp.~26--31, 2012.

\bibitem{adabound}
L.~Luo, Y.~Xiong, Y.~Liu, and X.~Sun, ``Adaptive gradient methods with dynamic
  bound of learning rate,'' {\em arXiv preprint arXiv:1902.09843}, 2019.

\bibitem{vsgd_fd}
T.~Schaul and Y.~LeCun, ``Adaptive learning rates and parallelization for
  stochastic, sparse, non-smooth gradients,'' {\em arXiv preprint
  arXiv:1301.3764}, 2013.

\bibitem{adasecant}
C.~Gulcehre, J.~Sotelo, M.~Moczulski, and Y.~Bengio, ``A robust adaptive
  stochastic gradient method for deep learning,'' in {\em 2017 International
  Joint Conference on Neural Networks (IJCNN)}, pp.~125--132, IEEE, 2017.

\bibitem{roadam}
Y.~Haimin, P.~Zhisong, and T.~Qing, ``Robust and adaptive online time series
  prediction with long short-term memory [j],'' {\em Computational Intelligence
  and Neuroscience}, vol.~2017, pp.~1--9, 2017.

\bibitem{holland2019efficient}
M.~J. Holland and K.~Ikeda, ``Efficient learning with robust gradient
  descent,'' {\em Machine Learning}, vol.~108, no.~8-9, pp.~1523--1560, 2019.

\bibitem{robustmeanestim}
M.~Lerasle and R.~I. Oliveira, ``Robust empirical mean estimators,'' {\em arXiv
  preprint arXiv:1112.3914}, 2011.

\bibitem{geometricmedian}
S.~Minsker {\em et~al.}, ``Geometric median and robust estimation in banach
  spaces,'' {\em Bernoulli}, vol.~21, no.~4, pp.~2308--2335, 2015.

\bibitem{riskmedian}
G.~Lugosi and S.~Mendelson, ``Risk minimization by median-of-means
  tournaments,'' {\em arXiv preprint arXiv:1608.00757}, 2016.

\bibitem{lossmedian}
D.~Hsu and S.~Sabato, ``Loss minimization and parameter estimation with heavy
  tails,'' {\em The Journal of Machine Learning Research}, vol.~17, no.~1,
  pp.~543--582, 2016.

\bibitem{mestim_riskminimization}
C.~Brownlees, E.~Joly, G.~Lugosi, {\em et~al.}, ``Empirical risk minimization
  for heavy-tailed losses,'' {\em The Annals of Statistics}, vol.~43, no.~6,
  pp.~2507--2536, 2015.

\bibitem{byzantineGD}
Y.~Chen, L.~Su, and J.~Xu, ``Distributed statistical machine learning in
  adversarial settings: Byzantine gradient descent,'' {\em ACM SIGMETRICS
  Performance Evaluation Review}, vol.~46, no.~1, pp.~96--96, 2019.

\bibitem{prasadrobustGD}
A.~Prasad, A.~S. Suggala, S.~Balakrishnan, and P.~Ravikumar, ``Robust
  estimation via robust gradient estimation,'' {\em arXiv preprint
  arXiv:1802.06485}, 2018.

\bibitem{robust_student_t1}
O.~Arslan, P.~D. Constable, and J.~T. Kent, ``Convergence behavior of the em
  algorithm for the multivariate t-distribution,'' {\em Communications in
  statistics-theory and methods}, vol.~24, no.~12, pp.~2981--3000, 1995.

\bibitem{robust_student_t2}
F.~Z. Do{\u{g}}ru, Y.~M. Bulut, and O.~Arslan, ``Doubly reweighted estimators
  for the parameters of the multivariate t-distribution,'' {\em Communications
  in Statistics-Theory and Methods}, vol.~47, no.~19, pp.~4751--4771, 2018.

\bibitem{reddi2019amsgrad}
S.~J. Reddi, S.~Kale, and S.~Kumar, ``On the convergence of adam and beyond,''
  {\em arXiv preprint arXiv:1904.09237}, 2019.

\bibitem{nettleton2010study}
D.~F. Nettleton, A.~Orriols-Puig, and A.~Fornells, ``A study of the effect of
  different types of noise on the precision of supervised learning
  techniques,'' {\em Artificial intelligence review}, vol.~33, no.~4,
  pp.~275--306, 2010.

\bibitem{relu}
W.~Shang, K.~Sohn, D.~Almeida, and H.~Lee, ``Understanding and improving
  convolutional neural networks via concatenated rectified linear units,'' in
  {\em international conference on machine learning}, pp.~2217--2225, 2016.

\bibitem{resnet34}
K.~He, X.~Zhang, S.~Ren, and J.~Sun, ``Deep residual learning for image
  recognition,'' in {\em Proceedings of the IEEE conference on computer vision
  and pattern recognition}, pp.~770--778, 2016.

\bibitem{coumans2019}
E.~Coumans and Y.~Bai, ``Pybullet, a python module for physics simulation for
  games, robotics and machine learning.'' \url{http://pybullet.org},
  2016--2019.

\bibitem{ppo}
J.~Schulman, F.~Wolski, P.~Dhariwal, A.~Radford, and O.~Klimov, ``Proximal
  policy optimization algorithms,'' {\em arXiv preprint arXiv:1707.06347},
  2017.

\bibitem{stooke2019rlpyt}
A.~Stooke and P.~Abbeel, ``rlpyt: A research code base for deep reinforcement
  learning in pytorch,'' {\em arXiv preprint arXiv:1909.01500}, 2019.

\bibitem{hartley1954unbiased}
H.~Hartley and A.~Ross, ``Unbiased ratio estimators,'' {\em Nature}, vol.~174,
  no.~4423, pp.~270--271, 1954.

\bibitem{goodman1958precision}
L.~A. Goodman and H.~O. Hartley, ``The precision of unbiased ratio-type
  estimators,'' {\em Journal of the American Statistical Association}, vol.~53,
  no.~282, pp.~491--508, 1958.

\end{thebibliography}

\end{document}